\documentclass[11pt]{article}

\usepackage{commath}
\usepackage[numbers]{natbib}
\usepackage[colorlinks=true,allcolors=blue]{hyperref}
\usepackage{amssymb}
\usepackage{makecell}
\usepackage{amsbsy}
\usepackage{amsmath}
\usepackage{amsthm}
\usepackage{amsfonts}
\usepackage{comment}
\usepackage{enumerate}
\usepackage{todonotes}
\usepackage{thmtools}

\usepackage{graphicx,bm,enumerate,fancyhdr,mathtools, multicol, amssymb, caption, adjustbox, multirow, mathabx}

\usepackage{mathrsfs,color,soul}
\usepackage{natbib}
\usepackage{commath}
\usepackage{longtable}
\usepackage{sectsty,subfigure}

\definecolor{darkred}{RGB}{150,50,50}
\definecolor{brown}{RGB}{250,100,100}
\definecolor{green}{RGB}{000,150,100}
\definecolor{purple}{RGB}{250,000,180}

\def\Psc{\bar{\Psc}}

\usepackage{xr}
\usepackage{hyperref}
\usepackage{verbatim}

\DeclareMathAlphabet{\mathpzc}{OT1}{pzc}{m}{it}
\hypersetup{
    colorlinks=true,
    citecolor=darkred,
    filecolor=red,
    linkcolor=red,
    urlcolor=red
}

\def\thick#1{\hbox{\rlap{$#1$}\kern0.25pt\rlap{$#1$}\kern0.25pt$#1$}}

\def\smbalpha{\boldsymbol{{\scriptstyle{\alpha}}}}

\def\what{{\widehat w}}

\def\smbalpha{\widehat{\smbalpha}}

\def\hbar{\bar{ h}}

\def\Psc{{\cal P}}

\def\E{\mbox{E}}

\def\mybox#1{\vskip1mm \begin{center}
        \hspace{.0\textwidth}\vbox{\hrule\hbox{\vrule\kern6pt
\parbox{.9\textwidth}{\kern6pt#1\vskip6pt}\kern6pt\vrule}\hrule}
        \end{center} \vskip-5mm}
\def\lboxit#1{\vbox{\hrule\hbox{\vrule\kern6pt
      \vbox{\kern6pt#1\vskip6pt}\kern6pt\vrule}\hrule}}

\def\thickboxit#1{\vbox{{\hrule height 1mm}\hbox{{\vrule width 1mm}\kern6pt
          \vbox{\kern6pt#1\kern6pt}\kern6pt{\vrule width 1mm}}
               {\hrule height 1mm}}}

\def\fat#1{\hbox{\rlap{$#1$}\kern0.25pt\rlap{$#1$}\kern0.25pt$#1$}}

\def\R{\mathbf{R}}
\def\what{\widehat}
\newcommand{\independent}{\perp\hspace{-.6em}\perp}

\newtheorem{theorem}{Theorem}

\newtheorem{assume}{Assumption}

\newtheorem{remark}{Remark}

\newcommand{\defeq}{\overset{\triangle}{=}}

\usepackage{setspace}

\usepackage[top=2.54cm,left=3cm,right=3cm,bottom=2.54cm]{geometry}

\begin{document}

\title{Estimation and Validation of Ratio-based Conditional Average Treatment Effects Using Observational Data}
\author{Steve Yadlowsky, Fabio Pellegrini, Federica Lionetto,  Stefan Braune, and Lu Tian}
  \begin{center}
  {\LARGE Estimation and Validation of Ratio-based Conditional Average Treatment Effects Using Observational Data} \\
  \vspace{.5cm} {\large Steve Yadlowsky$^1$, Fabio Pellegrini$^2$, Federica Lionetto$^3$,  Stefan Braune$^4$, and Lu Tian$^5$} \\
  \vspace{.2cm}
  \vspace{.2cm}
  $^1$Department of Electrical Engineering, Stanford University
  \\
  $^2$
  Global Head of RWE Innovations, ALIGN,  Biogen International GmbH
  \\
  $^3$PwC Data \& Analytics\\
  $^4$ Department of Neurology and Real World Evidence, NeuroTransData\\
  $^5$Department of Biomedical Data Science, Stanford University\\
  \vspace{.2cm}
  \texttt{syadlows@stanford.edu, fabio.pellegrini@biogen.com, federica.lionetto@gmail.com, sbraune@neurotransdata.com,\\ lutian@stanford.edu}
\end{center}

\begin{abstract}
\noindent
While sample sizes in randomized clinical trials are large enough to estimate the average treatment effect well, they are often insufficient for estimation of treatment-covariate interactions critical to studying data-driven precision medicine. Observational data from real world practice may play an important role in alleviating this problem. One common approach in trials is to predict the outcome of interest with separate regression models in each treatment arm, and estimate the treatment effect based on the contrast of the predictions. Unfortunately, this simple approach may induce spurious treatment-covariate interaction in observational studies when the regression model is misspecified.  Motivated by the need of modeling the number of relapses in multiple sclerosis patients, where the ratio of relapse rates is a natural choice of the treatment effect, we propose to estimate the conditional average treatment effect (CATE) as the ratio of expected potential outcomes, and derive a doubly robust estimator of this CATE in a semiparametric model of treatment-covariate interactions. We also provide a validation procedure to check the quality of the estimator on an independent sample. We conduct simulations to demonstrate the finite sample performance of the proposed methods, and illustrate their advantages on real data by examining the treatment effect of dimethyl fumarate compared to teriflunomide in multiple sclerosis patients.

\end{abstract}

\newpage
\section{Introduction}
Recently, interest in recommending tailored preventative interventions or treatments to patients in clinical practice has prompted investigating the conditional average treatment effect (CATE) from data. Knowledge of the CATE as the contrast between the expected outcome under different interventions conditional on covariate levels would allow clinicians to understand how mucuh a patient would benefit from a particular intervention based on their covariates. The primary statistical objective is to estimate these CATEs by examining treatment-covariate interactions \citep{tian2014simple}.

The small sample size of trials is one of the biggest obstacles in such analyses. Most randomized clinical trials are designed to study the average treatment effect (ATE), rather than the CATE.  Furthermore, to verify CATE estimates or high value subgroup of patients for whom the treatment is most effective, researchers use sample splitting \citep{athey2016recursive} or, ideally, independent external validation \citep{basu2017benefit} to account for the exploratory nature and the overfitting tendency of relevant statistical analyses, which further shrinks the available sample size. One important alternative is to use observational data from real world practice. Observational data often contain more samples, have broader target patient populations, and if collected from clinical practice, better represent realistic clinical conditions. However, patients receiving the treatment of interest and those receiving alternatives may be systematically different in observational data, which introduces new challenges in data analysis \citep{imbens2015causal}.

Let $Y^{(1)}$ be the outcome if the unit were given treatment $r=1$, and $Y^{(0)}$ be the outcome if given treatment $r=0$. In this work, we study the estimation and validation of the ratio-based CATE, 
$$D(z)=\frac{\E[Y^{(1)} | Z=z]}{\E[Y^{(0)} | Z=z]}$$
which targets the ratio of expected outcomes under different interventions.
Our motivation is the study of differences in the effect of treatments on relapse in multiple sclerosis (MS) patients, using observational data from the NeuroTransData registry (described below). For repeated events such as MS relapses, it's natural to look at the ratio of relapse rates under different potential treatments, as it is a relative measure of effectiveness.

We provide a general framework for the estimation and validation of such a ratio-based CATE score. We develop a doubly robust method for estimating the treatment contrast $D(z)$ in the semi-parametric model where $D(z) = \exp(\delta^\top z)$, but the conditional means, $\mu_{r}(z) = \E[Y^{(r)}\mid Z=z]$, and propensity score, $\pi_r(z) = P(R=r \mid Z=z)$, are in non-parametric models. This method has advantageous statistical properties, such as Neyman orthogonality \citep{chernozhukov2018double}, which, along with our simulation and experimental evidence, justify the use of machine learning methods for nuisance parameter estimation in the observational data. We provide another method for estimating the treatment contrast by adjusting per-arm regression analyses for confounding, and demonstrate via simulation that it often performs as well in practice. Both methods have the appealing property that when there is no treatment effect ratio heterogeneity, they will infer that $D(z)$ is constant, even if the regression model is mis-specified. Finally, we provide a method for validating the effectiveness of the learned CATE score in a statistically independent validation sample, and use this to compare the scores learned via different methods on the MS observational data.

\subsection{A motivating example}
\label{section:motivation}
The NeuroTransData (NTD) multiple sclerosis (MS) registry  includes about 25,000 patients with multiple sclerosis, which represents about 15\% of all MS patients in Germany. It includes demographic, clinical history, patient related outcomes and clinical variables captured in real time during clinical visits. The focus of the analysis is to estimate the CATE of Teriflunomide (TERI, $n= 1050$) compared with Dimethyl fumarate (DMF, $n= 1741$) and stratify the patient population for tailored treatment recommendation.
The outcome of the primary interest is the number of relapses per unit time; the average treatment effect is measured by the ratio of the expected relapse rate under TERI versus that under DMF. The ratio is important, because it better contextualizes the value of the treatment for these patients. Preventing 1 relapse per decade is more noticeable when the baseline is 1 relapse every 5 years than 1 relapse every year.

Three questions arise when trying to analyze CATEs in the NTD registry:
\begin{enumerate}
    \item Can we use the ratio rather than the difference in expected outcomes to measure the treatment effect?
    \item How do we adjust for differences in baseline covariates (confounders)
    between two treatment groups, and how do these affect estimation of the CATE?
    \item How do we validate our CATE model based on observational data and will the resulting method provide a useful measure of the quality of the estimated CATE score?
\end{enumerate}
The purpose of this paper is to answer these questions and provide a statistical methodology for estimation and validation of the ratio-based CATE, based on the described semiparametric model.

\subsection{Related Approaches}

The estimation of the absolute difference in expectations of potential outcomes as the CATE, i.e., $\E(Y^{(1)} - Y^{(0)} \mid Z = z),$ using observational data has been studied extensively in the literature \citep{green2012modeling, xie2012estimating, nie2017quasi, lu2018estimating, wager2018estimation, athey2019generalized, kunze2019metalearners}.  Recently, \cite{powers2018some} and \cite{wendling2018comparing} compare a number of approaches for learning this function. The basic idea is to either estimate the CATE based on separately estimated $\E(Y^{(r)}\mid Z=z), r=0, 1$, or learn directly using modified outcomes.  However, in some settings, the absolute difference in potential outcomes is not the best measure of treatment effect. For example, if the ratio-based CATE, $D(z),$ 
is constant for all $z$, but $\E[Y(0) | Z=z]$ varies with $z$, then there will appear to be significant treatment effect heterogeneity measured by the absolute difference $\E[Y^{(1)} - Y^{(0)} | Z=z],$ which may not be of particular interest. In this work, we focus on estimating the ratio-based CATE, i.e. the ratio of the conditional expectation of the potential outcomes given the baseline covariates. It's natural to consider applying the methodology for absolute differences to the $\log$-transformed outcomes. However, for counting processes where ratio-based measures are most natural, there is a nonzero probability that the observed outcome is zero, making $\E[\log(Y^{(r)}) \mid Z=z]$ infinite. However, our constrast $D(z)$ is well-defined whenever $\mu_0(z) > 0$.

In the context of binary outcomes, the ratio-based CATE (or ratio-based ATE) is known as the risk ratio.
\cite{robins2001inference} shows that
in a generalized linear model, the identity link function (corresponding to the absolute risk difference) and the log link function (corresponding to the risk ratio) are the only link functions that admit doubly robust estimators of the constant treatment effect. Given the extensive literature on CATE estimation with absolute differences, our work on ratio-based CATE estimation is a timely contribution. \cite{dukes2018note} study G-estimation of the risk ratio in the parametric setting, and demonstrate certain double robustness of the method.
\cite{van2011targeted} show that the related targeted maximum likelihood estimator for the risk ratio also satisfies double robustness.
In this paper, we demonstrate that the contrast regression posed is doubly robust in the discussed semi-parametric model and has the Neyman orthogonality property, which justifies the use of a wide class of non-parametric and machine learning methods for fitting the outcome and propensity score models in our method. Furthermore, the methodology and estimating equation used in \cite{van2011targeted} is optimized for binary outcomes, whereas our method is optimized for count data.

The outcome weighted learning (OWL) is another class of methods for developing precision medicine  \citep{zhao2012estimating, zhao2014doubly, chen2017general, zhou2017residual}. OWL methods find a decision boundary in the covariate space to classify patients into those with a treatment benefit $\{z\mid D(z)<1\}$ or harm $\{z\mid D(z)>1\},$ if $Y$ represents undesirable events.  The OWL method and its variations convert the original task into a binary classification problem and directly target the decision boundary, bypassing the need to estimate the CATE \citep{zhang2012estimating}. In contrast, the regression approach above attempts to directly estimate the magnitude of the benefit, and then identify the high value subgroup of patients accordingly \citep{cai2010analysis, foster2011subgroup, zhao2013effectively}. The OWL approach avoids the more difficult task, 
but also fails to yield information about the size of the treatment benefit for individual patient. 
A good estimator of the CATE ensures a good ATE within the subgroup consisting of patients with the largest CATEs, thus having priority of receiving the treatment. In addition, we may also directly recommend the treatment to patients whose estimated CATE outweighs the associated cost, which can also be patient-dependent. Therefore, in this paper, we focus on the more general question of directly estimating the CATE rather than a binary recommendation rule.

For data from a randomized clinical trial, \cite{zhao2013effectively} proposed the following approach to estimate and validate CATE. The method consists of two main steps:
\begin{enumerate}
   \item In the training set:
   \begin{itemize}
    \item Fit separate regression models $\what{\mu}_r(z) \approx \E[Y^{(r)}|R=r, Z=z]$ for the potential outcomes in the treatment ($r=1$) and control ($r=0$) arms.
    \item Estimate the CATE by  $\what D(z)=\what{\mu}_1(z)-\what{\mu}_0(z).$
    \end{itemize}
    \item In the validation set:
    \begin{itemize}
        \item Estimate $AD(c) = \E\left\{Y^{(1)} - Y^{(0)} \mid \what{D}(Z) \ge c\right\}$, the ATE for a subgroup of patients $\{z\mid \what D(z)\ge c\},$ and denote the resulting estimator by $\what{AD}(c)$.
        \item Draw the validation curve $q \mapsto  \what{AD}\{\what{H}^{-1}(1-q)\},$ where $q\in [0, 1)$ and $\what{H}(\cdot)$ is the empirical cumulative distribution function of $\what{D}(Z).$
        This curve graphically represents the relationship between the proportion of patients $q$ in the subgroup with the CATE score above $\what{H}^{-1}(1-q)$ and the estimated ATE in that subgroup. 
        \item Observe the slope of $\what{AD}\{\what{H}^{-1}(1-q)\}$, which reflects the quality of the scoring system $\what{D}(z)$ in ranking the patients according to their estimated CATE,  $\what{D}(z).$ 
    \end{itemize}
\end{enumerate}
In this paper, we extend the sequence of training and validation steps by \cite{zhao2013effectively} to the ratio-based CATE with observational data. 

\section{Method}

The standard regression model for the number of relapses in terms of the baseline covariates is the Poisson or Negative Binomial regression by treatment arm:
\begin{equation}
\E(Y^{(r)} \mid Z, R=r)=\exp(\beta_r^\top\widetilde{Z}),~r=0, 1,
\label{poisreg}
\end{equation}
\noindent
where $\widetilde{Z} = (1, Z^\top)^\top$ is a $d+1$ dimensional covariate vector, $R$ is the binary indicator of the treatment received, and $Y^{(r)}$ is the potential number of relapses if the patient received the treatment $r \in \{0,1\}$ or the ratio of the number relapses to an exposure time. We only observe $Y^{(r)}$ when $R=r$, but we are interested in $\E[Y^{(r)} \mid Z=z]$, the average outcome over all individuals with covariate $z,$ if they had been prescribed treatment $r$. This creates a causal-missing data problem. In this work, we make the unconfoundedness assumption \citep{imbens2015causal} that identifies the relapse rate:
\begin{equation}
    \{Y^{(1)}, Y^{(0)}\} \independent R \mid Z.
    \label{eq:unconfoundedness}
\end{equation}
This implies that 
$ \E(Y\mid Z, R=r)=\E(Y^{(r)} \mid Z, R=r) = \E(Y^{(r)} \mid Z).$

Our goal is to model the effect of the treatment on the relapse rate. We assume that the follow up time is the same for all patients. Otherwise, if unconfoundedness holds with respect to the follow up time, we may replace the outcome by the number of relapses divided by the exposure time.
Under the above model, the ratio of the expected relapse rates
\begin{equation}
\label{trueCATE}
D(z)=\frac{\E(Y^{(1)}|Z=z)}{\E(Y^{(0)}|Z=z)}
\end{equation}
is a natural measure of the CATE for the relapse rate that is insensitive to differences in exposure time between patients.
We can estimate $\beta_r$ by applying standard Poisson or Negative Binomial regression methods to the observed data in each arm to get the estimator $\widehat{\beta}_r$.
With the estimated regression coefficients $\what\beta_0$ and $\what\beta_1$, let
\begin{equation*}
    \what{\mu}_r(z) = \exp\{ \what{\beta}_r^\top \widetilde{z}\}
\end{equation*}
be an estimator of the relapse rate under treatment $r \in \{0,1\}$. A simple estimate of the CATE under model \eqref{poisreg} is
\begin{equation}
\what{D}(z)=\frac{\what{\mu}_1(z)}{\what{\mu}_0(z)}=\exp\{(\what\beta_1-\what\beta_0)^\top\widetilde{z}\}.
\label{itrprd}
\end{equation}

\subsection{Confounding effect on estimating the CATE}
\label{section:confounding-intro}
If the regression models for $Y$ given $Z$ and the treatment assignment $R$ are correctly specified,
then $\what\beta_r$ will converge to the true regression coefficient as the sample size increases regardless of the underlying distribution of $Z$ in each arm of the study. In practice, these statistical models may be mis-specified and only serve as working models approximating the true relationship between outcomes and covariates.  In such a case, the estimated regression coefficients may converge to limits that introduce spurious predicted treatment heterogeneity. 

The following toy example illustrates this phenomenon. Assume that $Y^{(r)} \mid Z=z$ follows a Poisson distribution with a rate of $z^2$ in both arms; therefore, $D(z) = 1$. To introduce confounding, assume that $Z \mid R=r \sim N\{(r-0.5), 1\}, r=0, 1.$  When fitting a mis-specified Poisson regression model 
$$\E(Y|Z=z, R=r)=\exp(\beta_r^\top\widetilde{z})$$ 
in two arms separately, the regression coefficient of $Z$ is $\beta_1=(-0.5, 0.8)^\top$ in arm $R=1$ and $\beta_0=(-0.5, -0.8)^\top$ in arm $R=0$. This is not surprising, because 70\% of the $Z_i$ in arm 1 are positive, where the quadratic function is increasing, inducing a positive association between $Y$ and $Z$, and 70\% of the $Z_i$ in arm 0 are negative in arm 0, where the quadratic function in decreasing, inducing a negative association.  Therefore, the estimated CATE score $\what{D}(z) = \exp\{(\what\beta_1-\what\beta_0)^\top\widetilde{z}\}\approx \exp(1.6 z)$ 
would suggest that the between group rate ratio increases with the value of $z,$ while in fact it is a constant.

This simple example shows that $\what{D}(z)$ estimated from misspecified regression models may create spurious treatment-covariate interactions in observational studies.
Therefore, the construction of the CATE score $\what{D}(z)$ should adjust for the covariates imbalance to avoid falsely demonstrating treatment effect heterogeneity due to confounding.

\subsection{Training}
In this section, we propose methodologies for estimating $D(z)$ using observational training data. First, we show how the parameters of the semi-parametric model specifying that $D(z) = \exp(\delta_0^\top \tilde{z})$ are identified, and use this to develop a doubly-robust method that can be used with machine learning estimates of the nuisance parameters and provide conditions that allow valid statistical inference. Then, we provide an approach based on fitting separate regressions for each treatment arm after adjusting for confounding effect.
This allows interpretation of the regression models in each arm , in the same way as fitting regression models to each arm of a randomized clinical trial. While the estimate of $\delta_0$ from the latter method may be biased, we show that it is consistent when there is no treatment heterogeneity.

\subsubsection{Contrast Regression Approach}
\label{section:contrast}
While motivated by the Poisson regression model, the CATE model
\begin{equation*}
    D(z) = \exp(\delta_0^\top \widetilde{z})
\end{equation*}
arises from a more general semi-parametric model that we will study in this section. Specifically, we model the conditional expectation of the potential outcomes $\mu_r(z)$ with the semi-parametric regression model
\begin{equation}
    \E(Y^{(r)}|Z=z) = \exp\left( r \delta_0^\top \widetilde{z} \right) \mu_0(z),
\label{eq:maineffect}
\end{equation}
where $\mu_0(z)$ is a unknown,  measurable and non-negative function in some non-parametric function class. We assume that the propensity score $\pi_r(z)$ is unknown, but is also in a non-parametric function class. This model represents the class of distributions for which $D(z)$ depends on $z$ through $\delta_0^\top z$, and thus our goal is to estimate $\delta_0$. We provide a doubly robust estimator of $\delta_0$ and discuss assumptions under which this estimator is $\sqrt{n}$-consistent.

If $Y^{(1)}$ and $Y^{(0)}$ were both observed, then $\delta_0$ is the solution to
\begin{equation}
\E\left[w(Z, \delta)\tilde{Z}\left\{Y^{(1)}-\exp(\delta^\top \tilde{Z}) Y^{(0)}\right\}\right]=0,
\label{eq:ideal-estimating-eq}
\end{equation}
because applying the tower property of conditional expectations to \eqref{eq:ideal-estimating-eq} gives the equivalent estimating equation $\E[w(Z, \delta)\widetilde{Z}\{ \mu_1(Z) - \exp(\delta^\top \widetilde{Z})\mu_0(Z)\}]=0,$ where $w(z, \delta)>0$ is a given weight function.
The solution is unique in any compact set $\Omega$ containing $\delta_0,$ as long as $\widetilde{Z}$ doesn't belong to a $d$ or lower dimensional hyperplane and $w(z, \delta)$ is bounded above and below, for all $\delta \in \Omega$ and $z$.  Like many causal inference and missing data problems, there are a variety of ways to develop estimating equations that are equivalent to \eqref{eq:ideal-estimating-eq} under condition \eqref{eq:unconfoundedness},
involving imputation with the mean functions $\mu_r(z)$ or inverse probability weighting using the propensity $\pi_r(z)$. 
Because the nuisance parameters $\mu_r(z)$ and $\pi_r(z)$ are rarely known in practice, operationalizing these estimators depends on estimating the nuisance parameters.

Therefore, we follow the approach advocated by \citet{robins2001inference} for developing doubly robust approaches for semi-parametric models. Our estimator is closely related to the generalized linear model with the logarithmic link function presented in their paper, and the doubly robust estimator of the semi-parametric risk ratio model presented in \cite{van2011targeted}. 
Specifically, for any candidate nuisance parameters $\mu : \R^d \to \R$ for the baseline mean (that will hopefully approximate $\mu_0$) and $\pi : \R^d \to [0, 1]$ for the propensity score (that will approximate $\pi_1(z)$) and parameters $\delta \in \Omega\subset\R^{d+1}$, we consider the estimating function
\begin{align*}
    m(G; \delta, \mu, \pi) &= \widetilde{Z} \frac{\{1-\pi(Z)\}RY - \pi(Z)(1-R)Y\exp(\delta^\top \widetilde{Z})}{ e^{\delta^\top \widetilde{Z}}\pi(Z) + (1 - \pi(Z))} \\
    &~~~~- \widetilde{Z}\mu(Z)\exp(\delta^\top \widetilde{Z}) \frac{R - \pi(Z)}{e^{\delta^\top \widetilde{Z}}\pi(Z) + (1 - \pi(Z))},
\end{align*}
where $G=(Y, R, Z^\top)^\top.$

If the propensity score is known, then substituting $\pi_1$ for $\pi$ gives the population estimating function
$$ \E\left[m(G; \delta, \mu, \pi_1) \right] =  \E\left[\widetilde{Z}w_1(Z; \delta, \pi_1)\left\{\mu_1(Z)-\mu_0(Z)e^{\delta^\top\widetilde{Z}}\right\}\right],$$
with the weight function
$$w_1(z; \delta, \pi)=\frac{\pi(z)(1-\pi(z))}{e^{\delta^\top\widetilde{z}}\pi(z)+1-\pi(z)},$$
and $\delta_0$ is a root of the corresponding estimating equation, for any bounded choice of $\mu$. 
In Section 1 of the Supplementary Materials, we show that this weight function is optimal in minimizing the variance of the resulting estimator when $Y^{(r)} \mid Z=z$ follows a Poisson distribution with rate $\mu_r(z)$ and with the correct propensity score model $\pi_1(z).$ 

On the other hand, by re-writing the estimating function as
\begin{align*}
m(G; \delta, \mu, \pi) &= \widetilde{Z} R\left\{Y-\mu(Z)\exp(\delta^\top\widetilde{Z})\right\}\frac{1-{\pi}(Z)}{e^{\delta^\top\widetilde{Z}}{\pi}(Z)+(1-{\pi}(Z))}\\ &~~~~+\widetilde{Z}(1-R)\left[Y-\mu(Z)\right]\frac{\exp(\delta^\top\widetilde{Z}){\pi}(Z)}{e^{\delta^\top\widetilde{Z}}{\pi}(Z)+(1-{\pi}(Z))},
\end{align*}
we observe that if $\mu_0(z)= \E(Y^{(r)}|Z=z)$ is known, then substituting $\mu_0$ for $\mu$ gives the population estimating equation
$$\E\left[ m(G; \delta, \mu_0, \pi) \right] = \E\left[\widetilde{Z} w_1(Z; \delta, \pi)\left\{\mu_1(Z)-\mu_0(Z)\exp(\delta^\top\widetilde{Z})\right\}\right]=0$$
for which $\delta_0$ is still a root, regardless of the choice of $\pi(z)$ used. 
In practice, neither the propensity score nor the conditional expectation $\mu_0(z)$ is known, which motivates the plug-in estimating equation
$$
S_n(\delta)= n^{-1}\sum_{i=1}^n m(G_i; \delta, \widehat{\mu}_0, \widehat{\pi}_1) = 0,
$$
where $G_i=(Y_i, R_i, Z_i^\top)^\top,$ $\widehat{\mu}_0(z)$ is an estimator for $\mu_0(z)$ and $\what{\pi}_1(z)$ is a estimator for the propensity score $\pi_1(z)$. If either $\what{\mu}_0(z)$ or $\what{\pi}_1(z)$ is consistent, then the solution of the estimating equation is a consistent estimator of $\delta_0$.

In the Appendix, we prove that this estimating equation satisfies the Neyman orthogonality condition \citep{chernozhukov2018double}. Therefore, using cross-fitting with this estimating equation allows for general use of machine learning estimators of the nuisance parameters, while still providing accurate confidence interval coverage. To compute the cross-fitting estimator $\widehat{\delta}_0$ of $\delta_0$, divide data into $K$ non-overlapping parts of approximately equal sizes indexed by ${\cal I}_k, k=1,\dots, K.$ Construct initial regression estimates
$\widehat{\mu}_r^{(-k)}(z)$ of $\mu_r(z)$ without using observations in ${\cal I}_k,$ and construct estimates $\what{\pi}_1^{(-k)}(z)$ of the propensity score $\pi_1(z)$ without using observations in ${\cal I}_k$ likewise. 
Then, estimate $\delta_0$ by searching for the root of the estimating equation
\begin{align*}
    S^{(cf)}_n(\delta)=& n^{-1} \sum_{k=1}^K\sum_{i\in {\cal I}_k} m(G_i; \delta, \widehat{\mu}_0^{(-k)}, \widehat{\pi}^{(-k)}_1).
\end{align*}

The estimator $\what{\delta}$ is $\sqrt{n}$-consistent and asymptotically normal under the following sufficient assumptions (Theorem~\ref{thm:semiparametric}).

\begin{assume}
(a) $Z \in \mathcal{Z}$, a bounded subset of $\R^d$ and the eigenvalues of $\E[ZZ^\top]$ are between $ \lambda_{\min} > 0$ and $\lambda_{\max}$,
(b) $\mu_1(z)$ and $\mu_0(z)$ are strictly positive and bounded on $\mathcal{Z}$, (c) There exists $\epsilon_\pi > 0$ such that $\epsilon_\pi \le \pi_1(z) \le 1-\epsilon_\pi$, (d) There exists $\sigma_L>0$ and $\sigma_U$ such that $\mbox{var}(Y^{(r)}\mid z)\in [\sigma_L, \sigma_U], z\in {\cal Z},$ (e) for some $q>2$, $\E[|Y^{(r)}|^q \mid Z=z]\le C < \infty$; (f) $\delta_0$ is a interior point of a compact set $\Omega\in \R^{d+1};$ and (g) $(\mu_0, \pi_1)\in {\cal T}$, which is a set of measurable functions and there exist positive constants $\epsilon_\pi$ and $\epsilon_\mu$ such that $\epsilon_\pi \le \pi(z) \le 1-\epsilon_\pi$, and $\epsilon_\mu \le  \mu(z) \le 1 / \epsilon_\mu $ for any $(\mu, \pi)\in {\cal T}.$ 
\label{assume:problem-regularity}
\end{assume}

\begin{assume}
There exists $n_0$, $\epsilon_\pi>0$, and $\epsilon_\mu>0$ such that for all $n > n_0$,
(a) $\epsilon_\pi \le \what{\pi}(z) \le 1-\epsilon_\pi$; (b) $\epsilon_\mu \le  \what{\mu}_0(z) \le 1 / \epsilon_\mu$;
(c) $\| \what{\mu}_0(z) - \mu_0(z) \|_{P,2}+\| \what{\pi}_1(z) - \pi_1(z) \|_{P,2} = o_P(n^{-1/4}).$ 
\label{assume:estimator-regularity}
\end{assume}

To specify the asymptotic distribution of the estimator $\what{\delta}$ under aforementioned assumptions, let
$$\what{w}^{(-k)}(\delta, Z_i)=\frac{e^{\delta^\top \widetilde{Z}_i}\what{\pi}_1^{(-k)}(Z_i)\widehat{\pi}_0^{(-k)}(Z_i)}{\left[e^{\delta^\top \widetilde{Z}_i}\what{\pi}_1^{(-k)}(Z_i)+\what{\pi}_0^{(-k)}(Z_i)\right]^2},$$
\begin{align*}
\what{A}(\delta)=&n^{-1}\sum_{k=1}^K\sum_{i \in {\cal I}_k}  \widetilde{Z}_i\widetilde{Z}_i^{\top} \what{w}^{(-k)}(\delta, Z_i)\left\{Y_i+\frac{\what{\mu}_0^{(-k)}(Z_i)}{\what{\pi}_1^{(-k)}(Z_i)}(R_i-\what{\pi}_1^{(-k)}(Z_i))\right\}\\
=&n^{-1}\sum_{k=1}^K\sum_{i \in {\cal I}_k}   \widetilde{Z}_i\widetilde{Z}_i^{\top} \what{w}^{(-k)}(\delta, Z_i)\left[R_iY_i+\frac{\what{\pi}_0^{(-k)}(Z_i)}{\what{\pi}_1^{(-k)}(Z_i)}\what{\mu}_0^{(-k)}R_i+(1-R_i)\{Y_i-\widehat{\mu}_0^{(-k)}(Z_i)\} \right],
\end{align*}
and
\begin{align*}
    \what{B}(\delta)=&n^{-1}\sum_{k=1}^K\sum_{i\in {\cal I}_k} \widetilde{Z}_i\widetilde{Z}_i^{\top}\left( R_i \frac{\left[Y_i-e^{\delta^\top \widetilde{Z}_i}\widehat{\mu}^{(-k)}_0(Z_i)\right]}{e^{\delta^\top \widetilde{Z}_i}\widehat{\pi}_1^{(-k)}(Z_i)+\widehat{\pi}_0^{(-k)}(Z_i)}\widehat{\pi}_0^{(-k)}(Z_i) \right.\\
    & \left.-(1-R_i)\frac{\left[Y_i-\widehat{\mu}^{(-k)}_0(Z_i)\right]e^{\delta^\top \widetilde{Z}_i}}{e^{\delta^\top \widetilde{Z}_i}\widehat{\pi}_1^{(-k)}(Z_i)+\widehat{\pi}_0^{(-k)}(Z_i)}\widehat{\pi}_1^{(-k)}(Z_i)\right)^{2}.
\end{align*}
Then, applying Theorem 3.3 from \cite{chernozhukov2018double} under Assumptions~\ref{assume:problem-regularity} and~\ref{assume:estimator-regularity} gives the following result:

\begin{theorem} Let $\what{\delta}$ solve $S^{(cf)}_n(\delta)=0$. Under Assumptions~\ref{assume:problem-regularity}~and~\ref{assume:estimator-regularity},
$\sqrt{n}(\widehat{\delta}-\delta_0)$ converges weakly to a mean zero Gaussian distribution, whose variance can be consistently estimated by 
$$\what{A}(\what{\delta})^{-1}\what{B}(\what{\delta})\what{A}(\what{\delta})^{-1}.$$
\label{thm:semiparametric}
\end{theorem}
See the Appendix for proof.  Assumption (\ref{assume:problem-regularity}) provides important regularity conditions that ensure finite, estimable parameters and nuisance parameters, and a non-degenerate asymptotic variance. Assumption (\ref{assume:estimator-regularity}) requires certain convergence rate for $\what{\mu}_0(\cdot)$ and $\what{\pi}_1(\cdot)$ in estimating $\mu_0(\cdot)$ and $\pi_1(\cdot),$ respectively. Under appropriate smoothness conditions for $\mu_0(z)$ and $\pi_1(z),$ there are multiple nonparametric estimators that achieve the required accuracy; see \citet{chernozhukov2018double} for a review of these estimators and their connection to cross-fitting estimators. The proposed estimating equation can be solved via Newton-Raphson method. Although we can't guarantee that the derivative matrix $\what{A}(\delta)$ is positive definite in finite samples, its limit is positive definite, with a consistent estimate of either the propensity score or the main effect $\mu_0(z)$. We find good numerical convergence in practice, when the sample size is adequately large. 

\begin{remark}
Constructing estimating equations $S_{n}^{(cf)}(\delta)$ based on different random partitions of the data and averaging the resulting solutions as the final estimator reduces the Monte-Carlo variation due to randomly splitting data into $K$ parts. \citet{chernozhukov2018double} shows that this estimator is asymptotically equivalent to $\what{\delta}$ analyzed above.
\end{remark}
\begin{remark}
\label{remark:symmetric}
By comparing the conditional means to the baseline of $\mu_1(z)$, the semi-parametric regression model (\ref{eq:maineffect}) is equivalent to 
$$\E \left[Y^{(r)} \mid Z=z \right] = \exp(-r \delta^\top \tilde{z})\mu_1(z),$$ 
and a similar analysis to the above gives a set of symmetric estimating equations in terms of nuisance parameters $\mu_1$ and $\pi_1$. In Section 2 of the Supplementary Materials, we provide procedure based on combining these symmetric estimating equations.

\end{remark}

\subsubsection{Two Regressions Approach}
\label{section:two-regressions}
Returning to the mis-specified regression from Section~\ref{section:confounding-intro}, recall that we were interested in ensuring that a CATE estimator will not introduce spurious heterogeneity due to confounding. If the data are from an RCT, so that $R \independent \{Z, Y^{(r)}\},$ then
$$\E\left[\widetilde{Z}_i\left\{\mu_r(Z_i)-\exp(\beta^\top\widetilde{Z}_i) \right\} \mid R=r \right] =  E\left[\widetilde{Z}_i\left\{\mu_r(Z_i)-\exp(\beta^\top\widetilde{Z}_i) \right\} \right]\defeq s_{r}(\beta).$$

Now, assume that there is no heterogeneity, so that \begin{equation} D(z)=\exp(d_0), \label{itrhomo}
\end{equation}
holds. Even though the regression may be mis-specified, the solutions $\beta_r^*$ of the estimating equations $s_r(\beta)=0$ in the two arms will satisfy $\beta_1^*-\beta_0^*=(d_0, 0, \cdots, 0)^\top,$ and thus, the estimated CATE score converges to $D(z)= \exp\{(\beta_1^*-\beta_0^*)^\top\widetilde{z}\}=\exp(d_0)$ in probability. This correctly suggests that there is no treatment effect heterogeneity even under mis-specified regression models. 

We propose an approach to correct for the confounding in observational data so that our estimator of the CATE will provide results as if estimated from a randomized trial, which avoids the spurious treatment effect heterogeneity from mis-specified regression models. Specifically, this approach ``recovers'' fitting a simple regression model in both arms of an RCT, while viewing the regression model as a working model approximating the association of interest, and fitting the regression model as if the potential outcomes and covariates are observed in the entire cohort.
Generally, this approach can produce biased estimate, when $\delta_0 \not= 0$ but model (\ref{eq:maineffect}) is correctly specified. However, in simulation and on the real data from the NTD registry, we find that it performs well compared to other methods.

Constructing empirical versions of the estimating equations $s_r(\beta)=0$ is not possible from the observed data, because $Y^{(r)}$ is only observed when $R=r$. Under the unconfoundedness assumption \eqref{eq:unconfoundedness},
we can apply methods developed to adjust for confounding when estimating the ATE to construct appropriate empirical estimating equations. To this end, let $\what\pi_r(z)$ be an estimator for the propensity score $\pi_r(z)=P(R=r|Z=z),$ and $$\what{W}_{i}(r)= r\frac{R_i}{\what\pi_1(Z_i)}+(1-r)\frac{1-R_i}{\what\pi_0(Z_i)}.$$
Then, we can use the doubly robust estimating equation
$$S_r(\beta)=n^{-1} \sum_{i=1}^n \widetilde{Z}_i\left\{\widetilde{\mu}_r(Z_i)-\exp(\beta_r^\top \widetilde{Z}_i)\right\}=0,$$
where $\widetilde{\mu}_r(z)$ is a special estimator of $\mu_r(z)$ constructed via the following steps:
\begin{enumerate}
    \item Construct an initial non-parametric (or otherwise more flexible parametric or semi-parametric) prediction for $Y_i^{(r)}$ given $Z_i=z$  via the estimated conditional expectation $\E(Y^{(r)}|Z=z),$ denoted by $\widehat{\mu}_r(z);$
    \item Solve the weighted estimating equations
    \begin{equation}n^{-1}\sum_{i=1}^n \what{W}_{i}(r) \widetilde{Z}_i\left(Y_i-\exp\left[\alpha_r\times\log\{\widehat{\mu}_r(Z_i)\}+\gamma_r^\top \widetilde{Z}_i\right]\right)=0, r=0, 1;\label{dr-step2}\end{equation}
    and denote the roots by $(\widehat{\alpha}_r, \widehat{\gamma}_r^\top)^\top, r=0, 1.$
    \item Let $\widetilde{\mu}_r(z)=\exp\left\{\widehat{\alpha}_r\times\log(\widehat{\mu}_r(Z_i))+\widehat{\gamma}_r^\top \widetilde{Z}_i\right\}$ be the ``calibrated" outcome predictions used in the estimating equation $S_r(\beta)=0.$
\end{enumerate}
This estimator is a doubly robust estimator: if either 
$\widehat{\mu}_r(\cdot)$ is a consistent estimator of $\mu_r(\cdot)$ or 
$\what{\pi}_r(\cdot)$ is a consistent estimator of $\pi_r(\cdot)$, then the solution to the augmented estimating equation converges to $\beta_r,$ the solution of $s_{r}(\beta)=0$ under  \eqref{eq:unconfoundedness} and mild regularity conditions \citep{bang2005doubly}.  The key observation is that if the propensity score is consistently estimated,  equation (\ref{dr-step2}) ensures that 
$$ \E\left[ \widetilde{Z} \{Y^{(r)}-\widetilde{\mu}_r(Z)\} \mid \widetilde{\mu}_r(\cdot)\right]=o_p(1).$$

If we suspect that the Poisson regression (\ref{poisreg}) is mis-specified, the initial prediction rule should be based a more flexible model than the Poisson regression that better approximates the true model. For example, we may fit a regression model 
$$\E(Y|Z, R=r)=\exp\{\eta_r^\top B(Z)\}, r=0,1,$$
where $B(z)$ is a rich set of basis functions capturing the complex nonlinear relationship between $Y$ and $Z.$ $\widehat{\mu}_r(z)=\exp\{\what\eta_r^\top B(z)\}$ can then be the initial prediction rule, where $\what{\eta}_r$ is the estimated regression coefficient.  Alternatively, we may employ machine learning methods such as random forest or boosting to generate $\what{\mu}_r(z)$ \citep{friedman2000additive, breiman2001random}.
We can improve the performance of the estimator by cross-fitting, which removes the dependence between $\what{\mu}_r(Z_i)$ and $(Y_i, Z_i)$ induced by potential overfitting in constructing $\what{\mu}_r(z).$ Specifically, $\what{\beta}_r$ is the solution to the estimating equation $S^{(cf)}_{r}(\beta) = 0,$ where
$$S^{(cf)}_{r}(\beta)=n^{-1} \sum_{k=1}^K \sum_{i \in {\cal I}_k} \widetilde{Z}_i\left\{\exp\left[\what{\alpha}_r\times\log\{\widehat{\mu}_r^{(-k)}(Z_i)\}+\what{\gamma}_r^\top \widetilde{Z}_i\right]-\exp(\beta_r^\top \widetilde{Z}_i)\right\},$$
 $\what{\gamma}_r$ and $\what{\alpha}_r$ are the roots of the estimating equation
$$ n^{-1}\sum_{k=1}^K \sum_{i \in {\cal I}_k} \what{W}_{i}^{(-k)}(r) \widetilde{Z}_i\left(Y_i-\exp\left[\alpha_r\times\log\{\widehat{\mu}_r^{(-k)}(Z_i)\}+\gamma_r^\top \widetilde{Z}_i\right]\right)=0,$$
data are divided into $K$ non-overlapping parts of approximately equal sizes indexed by ${\cal I}_k, k=1,\cdots, K,$ $\what{\mu}_r^{(-k)}(z)$ and $\what{\pi}_r^{(-k)}(z)$ are constructed using observations not in ${\cal I}_k,$ and $\what{W}_i^{(-k)}(r)$ is the analog of $\what{W}_i(r)$ with $\what{\pi}_r^{(-k)}(z)$ plugged in.
The estimated CATE score is thus 
    $$ \widehat{D}_1(z)=\exp\left\{(\widehat{\beta}_1-\widehat{\beta}_0)^\top \widetilde{z}\right\}.$$

\begin{remark} One natural question is that if $\widetilde{\mu}_r(z),$ a high quality prediction rule for $Y_i^{(r)}|Z_i=z,$ is already available, why do we need to reconstruct an estimator $\exp\{\what\beta_r^\top \widetilde{z}\}$ under a mis-specified regression model?
The initial prediction rule may be a complex function of $z;$ therefore, it is not as transparent as that based on a simple regression model for clinical interpretation and practical use.
We can view the regression-based CATE score as a ``projection" of the initial prediction $\widetilde{\mu}_r(z)$ to a simpler functional space.  This is in the same spirit of simplifying the estimated CATE by a classification tree  \citep{loh2015regression, foster2011subgroup}.
\end{remark}

\subsection{Validation}
\label{section:validation}
For estimating and validating CATE models, \cite{zhao2013effectively} considered the absolute difference, $\E(Y^{(1)}-Y^{(0)}\mid Z=z)$ and $\E(Y^{(1)}-Y^{(0)}| D(Z)\ge c).$ In such a case, the ATE in $\{z \mid D(z)\ge c\},$ is a monotone increasing function of $c.$ We generalize the method by \cite{zhao2013effectively} to address confounding due to differences in baseline covariates between two treatment groups, and validate CATE estimators of the ratio of expected potential outcomes.

Let the ratio of average treatment effects among the subgroup of patients with the highest true CATE $\{z : D(z) \ge c\}$ be
\begin{equation*}
    AD_{\text{true}}(c) = \frac{\E\{Y^{(1)} \mid D(z) \ge c\}}{\E\{Y^{(0)} \mid D(z) \ge c\}}.
\end{equation*}
Because this subgroup is selected based on the true treatment effect ratios, $AD_{\text{true}}(c)$ is monotone increasing in $c$. Instead, if we order patients by the estimated CATE score $\what{D}(z)$ and let
\begin{equation*}
    AD(c) = \frac{\E\{Y^{(1)} \mid \what{D}(z) \ge c\}}{\E\{Y^{(0)} \mid \what{D}(z) \ge c\}},
\end{equation*}
then we would expect that when $\what{D}(z)$ is a good estimate of $D(z)$, $AD(c)$ will be monotone increasing; the trend of $AD(c)$ is a natural measure of the quality of the CATE score. To estimate the ATE in the subgroup of patients $\{z\mid \what{D}(z)\ge c\}$, we show how to adjust for confounding between two treatment arms using propensity score, regression or doubly robust estimators \citep{bang2005doubly, kang2007demystifying}.

The following theorem ensures that for CATE measured by the ratio of 
$\mu_r(z)$, as in the NTD example, $AD(c)$ is also monotone increasing. 
\begin{theorem}
For nonnegative potential outcomes $Y^{(r)}, r=0, 1,$ Let 
$$D(z)=\frac{\E(Y^{(1)} \mid Z=z)}{\E(Y^{(0)}\mid Z=z)} $$
and 
$$AD(c)=\frac{\E(Y^{(1)}\mid D(z)\ge c)}{\E(Y^{(0)}\mid D(z)\ge c)}.$$
If all involved expectations are finite and $0 < D(Z) < \infty$ for almost every $Z$, $AD(c)$ is monotone increasing in $c,$ and $AD(c) \ge c$ for any $c$.
\label{thm:monotone-ad}
\end{theorem}
See Section 3 of the Supplementary Materials for detailed proof. By Theorem~\ref{thm:monotone-ad}, if we measure treatment effects by the ratio, we still can evaluate the quality of the constructed CATE scoring system by examining the ``slope'' of the curve $\what{AD}(c).$  Because, $AD(c)\ge c, $ for any $c,$ if $D(z)$ is the true CATE, the ATE in the subgroup consisting of patients with promising CATEs tends to be promising as well.

\begin{remark}
The monotonicity of $AD(c)$ depends on the metric used to measure the treatment effect. For example, $AD(c)$ is not necessarily monotone increasing if the treatment effect is measured by odds ratio (OR) for binary outcomes.
In Section 4 of the Supplementary Materials, we provide a simple example where the largest marginal OR is not in the subgroup of patients with highest conditional OR. Generally, the ATE in a subgroup of patients with the largest CATE may not be large if the treatment effect is measured by a contrast other than the ratio or difference. 
\end{remark}

To estimate $AD(c)$ using observational data, we need to account for potential imbalances in baseline covariates between two arms, since the treatment assignment is not randomized. There are various ways to estimate the ATE in an observational study and all involve certain model assumptions. To construct a doubly robust approach, suppose that the validation set consists of $m$, independent identically distributed copies of $(Y^V, R^V, Z^V),$ $\left\{(Y_i^V, R_i^V, Z_i^V), i=1, \cdots, m\right\},$ where the superscript $V$ indicates membership in the validation set. 
Then, estimate $AD(c)$ as follows: first estimate $\mu_r(z)$ by $\what{\mu}_{rc}(z)$ in the subgroup of patients $\{z^V \mid \hat{D}(z^V)\ge c\};$
and then estimate $\E(Y^{(r)} \mid \what{D}(Z^V)\ge c)$ by
$$\what\mu_r(c)=m_c^{-1}\sum_{\what{D}(Z_i^V)\ge c} \left[\what{\mu}_{cr}(Z_i^V)+\what{W}_{i}^V(r, c)\left\{Y_i^V-\what{\mu}_{cr}(Z_i^V)\right\}\right] ,$$
where
$$\what{W}_{i}^V(r, c)= r\frac{R_i^V}{\what{\pi}_{c1}(Z_i^V)}+(1-r)\frac{1-R_i^V}{\what\pi_{c0}(Z_i^V)},$$
 $\what{\pi}_{cr}(z)$ is the estimator for $\pi_r(z)$ in the subgroup $\{z^V \mid \what{D}(z^V)\ge c\}$ from the validation set, and $m_c$ is the subgroup size.
Finally, let $\what{AD}(c)$ be the simple plug-in estimator
$\what\mu_1(c)/\what\mu_0(c).$ This estimator is consistent for $AD(c)$, if either the propensity score or the main effect $\mu_r(z)$ is consistently estimated within the subgroup. One advantage of this approach is that it provides estimates of $\mu_r(z), r=0, 1,$ which allows interpretation of the treatment effect.

\begin{remark}
When the outcome of interest is time to a clinical event, such as death or relapse, the same method can be used to approximate and validate CATE, where the treatment effect is defined via the ratio of restricted mean time lost within a given time window, i.e, 
$$D(z)=\frac{\E\{\tau-(T^{(1)}\wedge \tau)|Z=z\}}{\E\{\tau-(T^{(0)}\wedge \tau) \mid Z=z\}}.$$
Here, $\tau>0$ is a chosen constant \citep{uno2014moving} and $T^{(j)}$ is the event time of interest under treatment $j$. When the event rate is low, this ratio is similar to the hazard ratio and can be approximated by a multiplicative model (\ref{eq:maineffect}). The detailed extension can be found in Section 5 of the Supplementary Materials.
\end{remark}

\section{Numerical Simulation} 
In this section, we conduct a simulation study to investigate the finite sample performance of the proposed method. The simulation design and discussion of the results are detailed below. Overall, these simulations show that when the ratio-based CATE is well-approximated by $D(z) = \exp(\delta^\top \widetilde{z})$ and the propensity score is correctly specified, the contrast regression and two regression approaches perform well. The contrast regression outperforms the two regression approach when the log-transformed CATE is well approximated by a linear function of $z$, but the Poisson model for the baseline rate $\mu_0(z)$ is mis-specified. 
When the log-transformed CATE is highly nonlinear, the increased flexibility of boosting and other machine learning methods is advantageous; the ratio of the predictions from boosting outperforms the proposed method.

In the simulations, the covariate $Z \in \R^{10}$ is generated from a multivariate Gaussian, where the first 5 components are independent and the last 5 components are correlated with a common correlation coefficient of 0.5 but independent of the first 5 components. The marginal distributions of the entries of $Z$ are standard Gaussians. To ensure that the propensity score is bounded away from 0 and 1, any $Z_i$ greater than 2 (or less than $-2$) is replaced by 2 (or $-2$). The treatment assignment $R \mid Z=z$ is generated from a Bernoulli distribution with a probability of $\pi_1(z)=\left\{1+\exp(z_1+0.5z_2-0.5z_6) \right\}^{-1}.$

We simulate a random follow up time $F^{(r)} \mid Z=z$ from a uniform distribution $U[0, 0.75]$ (discussion of incorporating follow up times is provided in the Supplementary Materials), and simulate the potential outcomes $Y^{(r)} \mid F^{(r)}=f, Z=z$ from a Poisson distribution $\mbox{Pois}\{\mu_r(z)f\}$, using the mean functions $\mu_r(z)$ described below, to illustrate a number of settings. With a slight abuse of notation, $z_i$ stands for the $i$-th component of the covariate vector rather than the covariate vector of the $i$-th patient in this section. The four different settings are:
\begin{enumerate}
    \item \textbf{Well-specified contrast}\vspace{-1ex}
      \begin{align*}
      D(z)=&\exp\left\{-0.1+0.25(z_1+z_6)\right\},\\
        \mu_1(z)=&\exp\left\{~0.85+0.25(z_1+z_6)+1.5(|z_1|-|z_6|)\right\}\\
        \mu_0(z)=&\exp\left\{~0.95~~~~~~~~~~~~~~~~~~~~+ 1.5(|z_1|-|z_6|)\right\};
    \end{align*}
     \item \textbf{Well-specified Poisson}\vspace{-1ex}
    \begin{align*}
    D(z) =& \exp \left\{0.375+0.125z_1+0.05z_2-0.25z_6 \right\}\\
        \mu_1(z)=&\exp\left\{0.925+0.125z_1+0.30z_2+0.25z_6\right\}\\
        \mu_0(z)=&\exp\left\{0.550~~~~~~~~~~~~~+0.25z_2+0.50z_6\right\};
    \end{align*}
    \item \textbf{Mild contrast mis-specification}\vspace{-1ex}
    \begin{align*}
    D(z)=&\exp\left\{~~0.75+0.125z_1+0.05|z_2+0.5|-0.25z_6 \right\}\\
        \mu_1(z)=&\exp\left\{~~0.50+0.125z_1+0.30|z_2+0.5|+0.25z_6+0.5(|z_1|+|z_6|)\right\}\\
        \mu_0(z)=&\exp\left\{ -0.25+~~~~~~~~~~~~~0.25|z_2+0.5|+0.50z_6+0.5(|z_1|+|z_6|)\right\};
    \end{align*}
    \item \textbf{Large contrast mis-specification}\vspace{-1ex}
    \begin{align*}
    D(z)=&\exp \left\{0.915-~0.25|z_1+z_6+1|-0.6|z_2+0.5|-0.25z_6 \right\} \\
        \mu_1(z)=&\exp\left\{1.235-0.125|z_1+z_6+1| -0.3|z_2+0.5|-0.125z_6+0.5(|z_1|+ |z_6|)\right\}\\
        \mu_0(z)=&\exp\left\{0.320+0.125|z_1+z_6+1| +0.3|z_2+0.5|+0.125 z_6+0.5(|z_1|+ |z_6|)\right\}.
    \end{align*}
\end{enumerate}
The proposed contrast regression is the most valuable for the well-specified contrast setting, where the Poisson regression is misspecified, but the underlying log-transformed CATE is still a linear combination of baseline covariates satisfying model (\ref{eq:maineffect}). We expect that the na\"ive regression approach should work especially well in the well-specified Poisson setting, but that the propensity adjusted two regression and contrast regression approaches should also perform reasonably well. 
In the remaining two settings, neither the Poisson model for $\mu_r(z)$ nor the semiparametric model (\ref{eq:maineffect}) for the CATE is correctly specified.  While the linear approximation is reasonably good in the third setting, the log-transformed CATE is highly nonlinear in the fourth setting.

For each simulated data set, we construct the CATE score using the following six methods:
\begin{enumerate}
\item \textbf{contrast regression} targeting $D(z)$ directly with the doubly robust adjustment,
\item \textbf{two regression} with the proposed doubly robust adjustment (the boosting-based estimation of $\mu_r(z)$ serves as the initial predictor),
\item \textbf{na\"ive regression} with a Poisson model in each arm,
\item \textbf{boosting} with a regression tree of depth 2 as the a base learner to estimate $\mu_r(z)$ in each arm separately and taking the ratio of two estimators,
\item \textbf{modified outcome (MO) boosting} regression according to \citet{wendling2018comparing},
\item \textbf{Bayesian additive regression tree (BART)} to estimate $\mu_r(z)$ and the CATE score taking the difference of two estimators \citep{lu2018estimating}.
\end{enumerate}
The propensity score, when used, is always estimated by fitting a standard logistic regression model. We calculated the true $AD\{\what H^{-1}(1-q)\}$ and the validation curve, $q \mapsto AD\{\what{H}^{-1}(q)\},$ based on the constructed CATE scores. The steeper the slope of the curve, the better the performance of the CATE score. We have also directly calculated the correlation coefficients between the estimated CATE score and the true CATE (after log-transformation).  After repeating this process 200 times, we summarize the performance of each method based on the median of the resulting validation curves (Figure \ref{fig:simVC}), where the validation curve of the true CATE serves as the benchmark. Figure \ref{fig:simCC} shows the distribution of the correlation coefficients between the estimated CATE score and the truth. As expected, the contrast regression outperforms the two regression and other approaches in ranking the magnitude of CATE in the well-specified contrast setting. In most other cases, the CATE estimated by the two proposed methods have similar concordance with the true CATE. In the simulation for the well-specified Poisson model, the na\"ive regression performs the best, however the boosting and two proposed methods are only slightly inferior to the na\"ive regression. In the large contrast mis-specification setting, where the log-transformed CATE is highly nonlinear, the ratio of the predictions from boosting outperforms the proposed methods, suggesting that the increased flexibility of the boosting approach (or other machine learning method) is advantageous.

In addition, for settings 1 and 2, model \ref{eq:maineffect} is correctly specified and $\what{\delta}$ from the contrast regression is a consistent estimator of $\delta_0.$ We have also examined the empirical bias and the coverage level of 95\% confidence intervals in estimating $\delta_0$ based on 400 replications. The results are summarized in Table \ref{table:simulation}.
The proposed contrast regression estimator is almost unbiased and the empirical coverage level of the constructed 95\% confidence interval is close to the nominal level. 

We also designed some simulations for survival outcomes, which can be found in Section 7 of the Supplementary Materials. In general, the methods provided here adapt well to survival outcomes with censoring.

\section{Treatments for Multiple Sclerosis} 
We return to our motivating example--measuring treatment effect heterogeneity between the TERI and DMF drugs for MS. As discussed in Section~\ref{section:motivation}, one of the primary endpoints of interest is the relapse rate of severe symptoms of MS. The NTD registry records observational data of MS patients, including their treatments, relapses, and covariates over time. Hypothesized heterogeneity may be due to different drug treatment pathways, leading to different effectiveness across individuals. Here, we provide an in-depth description of the data, the results of applying the proposed analyses on these observational data, and some implications of using the proposed methods to measure the CATE.  Additional analysis and results for the time to relapse can be found in Section 8 of the Supplementary Materials.

\subsection{Experimental design}
The NTD registry captured 1050 MS patients receiving TERI and 1741 patients receiving DMF between January 1, 2009 and July 1, 2018. Covariates of interest include age, number of prior treatments, MS duration, prior usage of glatiramer acetate (GA), prior usage of interferon (IFN), number of relapses in one year and in two years prior to the index therapy, baseline Expanded Disability Status Scale (EDSS),  and baseline pyramidal EDSS score. The data contain few missing values, thanks to processes to manage the definition of minimum data sets, mandatory data entry fields, and positive missing data confirmation. More details on the data source and management are available in Section 9 of the Supplementary Materials.

We implemented the standard regression, and the two proposed methods to construct the CATE scores approximating the ``individualized" relapse rate ratio. To implement the proposed procedure, we estimated the baseline relapse rates $\mu_r(z)$ using boosting with the Poisson likelihood. The base learners are depth 2 regression trees, and the number of trees is selected via 5-fold cross-validation. The propensity score is constructed based on the standard logistic regression model. The proposed CATE score is based on the average of three replicates of 7-fold cross-fitting. One advantage of the contrast regression is that the standard errors can be estimated using the formulas provided in Section~\ref{section:contrast}, and so we include these in our results.

We use repeated cross validation to compare and evaluate the performance of the CATE scores objectively using the validation curves described in Section~\ref{section:validation}.
To this end, we considered four CATE scores: a score based on predicted relapse rates using boosting method, a score based on na\"ive Poisson regression, and two scores based on our new proposal with the boosting-based prediction as the initial prediction.  In each iteration, the data are split into a training set (67\%) used to fit the CATE score and a testing set (33\%) used to construct the validation curve. After repeating this process 50 times, we report thetwise median validation curve for each CATE score in the training (left) and test (right) sets.
We also use the estimated CATE score to split the patients in the testing set into two group of equal sizes. Then, we estimate the ratio of average relapse rates in two subgroups separately.

\subsection{Results}
Table \ref{table:baseline} summarizes the distribution of covariates by treatment arm. The patients receiving TERI are different from patients receiving DMF in several key ways. For example, the patients receiving TERI tend to be older (45 vs 40), have a longer disease duration (8.1 years vs. 6.6 years) and have higher EDSS scores (2.03 vs 1.84) than those receiving DMF.

In the entire cohort, the estimated ratio of the relapse rates (TERI vs DMF) is 1.270 (95\% confidence interval (CI): 1.121, 1.439; $p<0.001)$ using Poisson regression alone.  After adjusting for confounding using the doubly robust procedure, the annual relapse rate is 0.308 for TERI and 0.237 for DMF; the estimated relapse rate ratio is 1.299 (95\% CI 1.018, 1.658).

Table \ref{table:score} summarizes the estimated weights in the log-transformed CATE score for the na\"ive regression, two regression, and contrast regression approaches, as well as the estimated standard errors from the contrast regression. Based on contrast regression, GA, the number of relapses in the year prior to the therapy, the number of relapses in two years prior to the therapy, and baseline EDSS have a statistically significant impact on the treatment effectiveness at the 0.05 level. These weights suggest, for example, that patients experiencing more relapses in the previous two years and has a lower EDSS score tend to benefit more from DMF, even when looking at the relative ratio of treatment benefit.

The composition of the CATE score based on the na\"ive Poisson regression is different from that based on two new proposals. For example, the weight of EDSS from the na\"ive approach is substantially smaller than those in new CATE scores. Figure~\ref{fig:cor} shows a scatter plot of these three CATE scores in the entire cohort, demonstrating a positive correlation but also ample differences between the na\"ive and new CATE scores.  On the other hand, the two new CATE scores are highly concordant. Comparing the cross-validation performance in Figure~\ref{fig:exp1}, the two proposed CATE scores that adjust for imbalance in baseline covariates appear to have a similarly superior performance to na\"ive regression and boosting in the testing set, and both suggest a moderate treatment effect heterogeneity due to the monotone shape (with non-monotone noise) of the validation curve.

Using the CATE score from contrast regression to split the patients in the testing set into two equal groups, the median ratio of the relapse rates (TERI vs DMF) is 1.673 in 50\% patients that would benefit most from DMF and 1.113 in remaining 50\% patients. Using the CATE score from the two regression approach gives median ratios of 1.723 and 1.089, respectively. The distribution of the estimated ratios across different cross-validation replicates are summarized in Figure \ref{fig:exp2}.
 
As a cautionary note, this observed difference in treatment effect may not be adequately stable due to the limited sample size in the testing set (on average, there are only 465 patients in each of the two subgroups). However, the results still exhibit signals for the presence of treatment effect heterogeneity captured by two proposed approaches.

A important observation is that the estimated treatment effect heterogeneity doesn't alter the recommendation of the treatment, since DMF appears to be superior to TERI in most if not all of the patients in terms of reducing relapse rate, although the relative benefit may vary in different subgroups.

\section{Discussion}
We show that estimation and validation of the ratio-based CATE benefits from many of the same approaches such as doubly robust estimation and semi-parametric modeling that work well for the difference-based CATE. We also extend the regression approach by \cite{zhao2013effectively} to develop a precision medicine strategy from observational data. There are three important messages learned in this practice. First, the metric for the treatment effect has an important impact on the estimation and validation of the CATE. The treatment effects measured by the absolute difference and relative ratio both depend on the outcomes distribution in the control arm in simple ways, such that that the group of patients with large CATEs also have a large ATE. This is not necessarily true for treatment effect measured by odds or hazard ratio, where the ATE and CATE don't always align.
Second, by borrowing appropriate techniques developed for estimating ATE in causal inference to adjust the standard estimation procedure, we eliminate the spurious heterogeneity caused by the imbalance in covariates in regression modeling for treatment effect interactions. Lastly, we proposed a set of methods for estimating the ratio-based CATE, which may result in very different conclusions in comparison with most current methods, which target difference-based CATE.

We note that in this work, we have assumed that training and validation sets follow the same distribution. If the distribution of the validation set or the target population is different from that of the training set, the proposed estimation procedures need to be modified to adjust the distribution of covariates of the patients in treatment arm $r$ of the training set to match that of the target population. Otherwise, the same CATE score may define a different subgroup of patients in the target population, i.e., $\{z\mid \what{D}(z)\ge c\}$ may be different from $\{z^V \mid \what{D}(z^V)\ge c\}$, so that the ATE observed in the high value subgroup $\{z\mid \what{D}(z)\ge c\}$ may not be reproducible. Furthermore, the validity of all of our results depends on making the unconfoundedness assumption for causal effects.
This is the most important for CATE validation, which would benefit greatly from being an external randomized clinical trial, so that we can estimate the ATE in the identified high value subgroup without systematic biases from unmeasured confounding.

\cite{qi2019multi} discussed the optimal treatment recommendation in the presence of $K>2$ treatments. The proposed two regression approach can be used to approximate $\mu_k(z)=\E(Y^{(r)}\mid Z=z), k=0, 1, \cdots, K,$ and select treatments accordingly. The contrast regression can directly estimate $D_{ij}(z)=\mu_i(z)/\mu_j(z)$ based on the limiting estimating equation:
$$ \E \left[w(Z, \delta)\widetilde{Z}\left\{ \prod_{k\neq i}\pi_k(Z)I(R=i)Y-\prod_{k\neq j}\pi_k(Z)I(R=j)Y\exp(\delta^\top \widetilde{Z})\right\} \right]=0$$
for estimating $D_{ij}(z)=\exp(\delta_{ij}^\top \widetilde{z}),$ where $\pi_k(z)=P(R=k\mid Z=z)$ and $w(z, \delta_{ij})$ is a weight function. Appropriate doubly robust augmentation based on  $\{(I(R=r)-\pi_r(Z),  r=1,\cdots, K\}$ may further improve efficiency. However, the resulting estimators don't necessarily have the property that $D_{ij}(z)=D_{il}(z)D_{lj}(z),$ which warrants further research.

\subsection*{Acknowledgements}
The authors thank Hongseok Namkoong and the anonymous reviewers for helpful comments. SY is partially supported by the Stanford Graduate Fellowship and NHLBI award R01HL144555-01. Dr.~Tian's research is partially supported by R01HL089778-05 and
\\1UL1TR003142
from National Institutes of Health, USA.

\bibliographystyle{abbrvnat}      
\bibliography{SSLref.bib}

\appendix
\section{Appendix: Proof of Theorem 1}
\begin{proof}
To prove Theorem 1, it is sufficient to verify that the problem and assumptions satisfy those of Theorem 3.3 in \cite{chernozhukov2018double},
which we repeat here for the reader's convenience. Let $c_0 > 0$, $c_1>0$, $a > 1$, $v > 0$, $s > 0$, and $q>2$ be finite constants, and let $\{\delta_n\}_{n \ge 1}$, $\{\Delta_n\}_{n \ge 1}$, and $\{\tau_n\}_{n \ge 1}$ be some sequence of positive constants converging to $0$.
Define the following assumptions \cite[Assumptions 3.3~and~3.4]{chernozhukov2018double}.
\begin{assume}
For all $n \ge 3$ and $P \in \mathcal{P}_n$,
(a) $\E\{m(G;\delta_0, \mu_0, \pi_1)\}=0$, and $\Omega$ contains a ball of radius $c_1 n^{-1/2}\log n$ centered at $\delta_0;$
(b) the map $(\delta, \mu, \pi) \rightarrow \E\{m(G; \delta, \mu, \pi)\} $ is twice continuously Gateaux-differentiable on $\Omega \times {\cal T};$ (c) for all $\delta \in \Omega,$ $2\|\E\{m(G;\delta, \mu_0, \pi_1)\}\|\ge \|J_0(\delta-\delta_0)\|\wedge c_0,$ where $J_0$ is the Jacobian matrix of $\delta \mapsto \E[m(G; \delta, \mu_0, \pi_1]$ at $\delta_0;$ (d) the score $m(g; \delta, \mu, \pi)$ obeys the Neyman orthogonality condition
\begin{equation*}
    \frac{\dif{}}{\dif{r}} \E\left[ m(G, \delta_0, \mu_0 + r(\widebar{\mu} - \mu_0), \pi_1 + r(\widebar{\pi}-\pi_1)) \right] \bigg{|}_{r=0} = 0,
\end{equation*}
for any $(\widebar{\mu}, \widebar{\pi}) \in {\cal T}.$
\label{assume:dml-problem-reg}
\end{assume}

\begin{assume}
    Let $K$ be a fixed integer. For all $n \ge 3$ and $P \in \mathcal{P}_n$, the following conditions hold:
    (a) Given a random subset $I$ of $\{1,\dots,n\}$ of size $n/K$, the nuisance parameter estimators $(\what{\mu}_0^{-k}, \what{\pi}_1^{-k})_{1 \le k \le K}$ belong to the realization set $\mathcal{T}_n$ with probability $1 - \Delta_n$, where $\mathcal{T}_n$ contains $(\mu_0, \pi_1)$ and is constrained by the conditions below; (b) ${\cal F}_{1, (\mu, \pi)}=\{m_j(g; \delta, \mu, \pi) \mid j=1, \cdots, d+1, \delta_0\in  \Omega\}$ is suitably measurable and its uniform covering entropy obeys
    $$\sup_{Q}\log N(\epsilon\|F_{1,(\mu, \pi)}\|_{Q,2}, {\cal F}_{1, (\mu, \pi)}, \|\cdot\|_{Q,2})\le v\log(a/\epsilon_N)$$
    for $\epsilon_N \in (0, 1]$, where $F_{1, (\mu,\pi)}$ is a measurable envelope for $\mathcal{F}_{1,(\mu, \pi)}$ that satisfies $\| F_{1,(\mu,\pi)} \|_{P,q} \le c_1;$
    (c) $r_n=\sup_{(\mu, \pi)\in {\cal T}_n, \delta_0 \in \Omega} \| \E\{m(G; \delta, \mu, \pi)\}-\E\{m(G; \delta_0, \mu_0, \pi_0)\}\|\le \delta_n\tau_n;$
    (d) $r_n'\log^{1/2}(1/r'_n)\le \delta_n,$ where
    $$r_n'=\sup_{(\mu, \pi)\in {\cal T}_n, \|\delta-\delta_0\|\le\tau_n}  \left(\E\left\{\|m(G; \delta, \mu, \pi)-m(G; \delta_0, \mu_0, \pi_0)\|^2\right\}\right)^{1/2};$$
    (e) $\lambda_n\le \delta_n n^{-1/2},$ where
    $$\lambda_n=\sup_{r\in (0,1), (\mu, \pi)\in {\cal T}_n, \|\delta-\delta_0\|\le \tau_n}\| \partial^2_r \E \left[m\{G; \delta_0+r(\delta-\delta_0), \mu_0+r(\mu-\mu_0), \pi_1+r(\pi-\pi_1)\}\right]\|.$$
    (f) all eigenvalues of the matrix $ \E\left[m(G; \delta_0, \mu_0, \pi_1)m^\top (G; \delta_0, \mu_0, \pi_1)\right]$ are bounded below by a positive constant.
\label{assume:dml-est-reg}
\end{assume}

\begin{theorem}[{\cite[Theorem 3.3]{chernozhukov2018double}}] Suppose that Assumptions~\ref{assume:dml-problem-reg} and \ref{assume:dml-est-reg} hold. In addition, suppose that $\delta_n \ge n^{-1/2 + 1/q} \log(n)$ and that $n^{-1/2}\log(n) \le \tau_n \le \delta_n$ for all $n \ge 1$ and a constant $q>2.$ Then, the DML2 estimator $\widehat{\delta}$
concentrates in a $1/\sqrt{n}$ neighborhood of $\delta_0$, and are approximately linear and centered Gaussian:
\begin{align*}
    \sqrt{n}\sigma^{-1}\left( \widehat{\delta} - \delta\right) = \frac{1}{\sqrt{n}} \sum_{i=1}^n \bar{\psi}(G_i) + O_p(\rho_n) \overset{d}{\to} N(0,I),
\end{align*}
uniformly over $P \in \mathcal{P}_n$, where the size of the remainder term obeys
\begin{equation*}
    \rho_n = n^{-1/2 + 1/q} \log(n) + r_n' \log^{1/2}(1/r_n') + n^{1/2}\lambda_n + n^{1/2}\lambda_n',
\end{equation*}
$\bar{\psi}(\cdot) = -\sigma^{-1}J_0^{-1}m(\cdot, \delta_0, \mu_0, \pi_1)$ is the influence function, and the approximate variance is
\begin{equation*}
    \Sigma = J_0^{-1} \E\left[ m(G, \delta_0, \mu_0, \pi_1)m(G, \delta_0, \mu_0, \pi_1)^{\top} \right] J_0^{-\top}.
\end{equation*}
\label{thm:chernozhukov}
\end{theorem}

We proceed by verifying the assumptions of Theorem~\ref{thm:chernozhukov}.
Let $\mathcal{P}$ be a set of measures satisfying Assumption~\ref{assume:problem-regularity} and ${\cal T}$ be a measurable subset of the pairs of functions $(\pi,\mu)$
such that for each $Q \in \mathcal{P}$,
$\mu \in L_2(Q)$, $\pi \in L_{\infty}(Q)$,
and $\epsilon_\pi \le \pi(z) \le 1 - \epsilon_\pi$
$Q$-almost everywhere. 

We proved Assumption~\ref{assume:dml-problem-reg}(a) in the main text; see equation \eqref{eq:ideal-estimating-eq}, by using the assumption in \eqref{eq:maineffect}: $\mu_1(z) = \mu_0(z)\exp(\delta_0^\top \widetilde{z})$, and thus 
$$\E[m(G; \delta_0, \mu_0)] = \E[w(Z, \delta_0)(\mu_1(Z) - \mu_0(Z)\exp(\delta_0^\top \widetilde{Z}))] =0.$$

We will frequently use the fact that for $\delta \in \Omega$, $Z \in \mathcal{Z}$, and $\pi(z)$ satisfying $0 < \epsilon_\pi \le \pi(z) \le 1- \epsilon_\pi$, 
\begin{equation}
   \sup_{z\in {\cal Z}, \delta\in \Omega} \{\exp(\delta^\top \widetilde{z})\pi(z) + 1-\pi(z)\}^{-1}\le C_0
    \label{eq:bounded-denom}
\end{equation}
for a constant $C_0.$ Applying this with $\pi = \pi_1$ by Assumption~\ref{assume:problem-regularity}(c), along with Assumption~\ref{assume:problem-regularity}(e) ensures that $\partial \E[m(G; \delta, \mu_0, \pi_1) \mid Z=z]/\partial \delta$ and the second derivative have an integrable envelope function, and therefore
$\E\left[m(G; \delta, \mu_0, \pi_1)\right]$ is differentiable with respect to $\delta$ with Jacobian 
$$ J_0(\delta)=\E \left( \widetilde{Z}\widetilde{Z}^\top \frac{e^{\delta^\top \widetilde{Z}_i}\pi_1(Z)\pi_0(Z)\{e^{\delta_0^\top \widetilde{Z}}\pi_1(Z)+\pi_0(Z)\}\mu_0(Z)}{\left[e^{\delta^\top \widetilde{Z}}\pi_1(Z)+\pi_0(Z)\right]^2} \right),$$
which is continuous in $\delta$ and positive definite with its smallest eigenvalue uniformly bounded away from zero for $\delta\in \Omega.$ We can choose a small open ball centered at $\delta_0$, ${\cal N},$ such that for any $\delta\in {\cal N}, |J_0(\delta)-J_0(\delta_0)|_{ij}<\epsilon,$ for all components $1\le i, j\le d+1$, where $\epsilon$ is a small constant to be specified later.  By the Intermediate Value Theorem, for any $\delta\in {\cal N},$ there exists $\bar{\delta}\in {\cal N}$ such that 
\begin{align*}
    &\|\E\{m(G;\delta, \mu_0, \pi_1)\}\|=\|J_0(\bar{\delta})(\delta-\delta_0)\|\ge  \|J_0(\delta_0)(\delta-\delta_0)\|-\|\{J_0(\bar{\delta})-J_0(\delta_0)\}(\delta-\delta_0)\|\\
\ge & \|J_0(\delta_0)(\delta-\delta_0)\|-\epsilon(d+1)\|\delta-\delta_0\|\ge  \|J_0(\delta_0)(\delta-\delta_0)\|-\|J_0(\delta_0)(\delta-\delta_0)\|/2,
\end{align*}
if $\epsilon\le \lambda_0/2(d+1),$ where $\lambda_0$ is the smallest eigenvalue of 
$$J_0=J_0(\delta_0)=\E \left( \widetilde{Z}\widetilde{Z}^\top \frac{e^{\delta_0^\top \widetilde{Z}_i}\pi_1(Z)\pi_0(Z)\mu_0(Z)}{\left[e^{\delta_0^\top \widetilde{Z}}\pi_1(Z)+\pi_0(Z)\right]} \right).$$
Therefore, $2\|\E\{m(G;\delta, \mu_0, \pi_1)\}\|\ge \|J_0(\delta-\delta_0)\|,$
for any $\delta$ within the ball.  For $\delta$ outside ${\cal N},$ let 
$c_0=\inf_{\delta \in \Omega-{\cal N}}\|\E\{m(G;\delta, \mu_0, \pi_1)\}\|.$
$c_0>0$ due to the uniform continuity of $\E\{m(G;\delta, \mu_0, \pi_1)\}$ in the compact set $\Omega-{\cal N}.$ This verifies Assumption~\ref{assume:dml-problem-reg}(c).

To verify the Gateaux-differentiability of  $m$, note that for $\delta \in \Omega,$ $(\mu, \pi)\in {\cal T},$ $(\mu+rd_\mu, \pi)\in {\cal T},$ and $(\mu, \pi+rd_\pi)\in {\cal T},$
\begin{align*}
&\frac{1}{r}\left[\E\{m(G; \delta, \mu+r d_\mu, \pi)\}- \E\{m(G; \delta, \mu, \pi)\}\right] 
= -\E\left\{\frac{ \left\{\pi_1(Z)(1-\pi(Z)) -\pi_0(Z)\pi(Z) \right\} e^{\delta^\top\widetilde{Z}} }{e^{\delta^\top\widetilde{Z}}\pi(Z)+1-\pi(Z)} d_\mu(Z) \right\},
\end{align*}
and 
\begin{align*}
&\frac{1}{r}\left[\E\{m(G; \delta, \mu, \pi+rd_\pi)\}- \E\{m(G; \delta,  \mu, \pi)\}\right]\\   
=& -\E\left\{\frac{ \left[\pi_1(Z)\{\mu_1(Z)-\mu(Z)e^{\delta^\top\widetilde{Z}}\}+\pi_0(Z)\{\mu_0(Z)-\mu(Z)\} \right]  e^{\delta^\top\widetilde{Z}} }{\left(e^{\delta^\top\widetilde{Z}}\pi(Z)+1-\pi(Z)\right)\left(e^{\delta^\top\widetilde{Z}}\{\pi(Z)+rd_\pi(Z)\}+1-\pi(Z)-rd_\pi(Z)\right)}d_\pi(Z) \right\}.
\end{align*}
By Dominated Convergence Theorem,  we may exchange the $\lim_{r\rightarrow 0}$ with the expectation, and 
the Gateuaux derivative with respect to $\mu$ along the direction of $d_\mu$ exists. 
Similarly, the Gateuaux derivative with respect to $\pi$ along the direction of $d_\pi$ also exists, and is 
$$-\E\left\{\frac{ \left[\pi_1(Z)\{\mu_1(Z)-\mu(Z)e^{\delta^\top\widetilde{Z}}\}+\pi_0(Z)\{\mu_0(Z)-\mu(Z)\} \right]  e^{\delta^\top\widetilde{Z}} }{\left\{e^{\delta^\top\widetilde{Z}}\pi(Z)+1-\pi(Z)\right\}^2}d_\pi(Z) \right\}.$$
The smoothness of the numerator, and boundedness of the denominator similarly allow for second order differentiability.

To examine the orthogonality condition, i.e., Assumption~\ref{assume:dml-problem-reg}(d), let $(\widebar{\mu}, \widebar{\pi}) \in {\cal T},$ $d_\mu(z) = \widebar{\mu}(z) - \mu_0(z)$, $d_\pi(z) = \widebar{\pi}(z) - \pi_1(z)$, and
$$\bar{m}(\mu, \pi, f)=\E\left[Z\left\{\frac{(1-\pi(Z))R(Y-\mu(Z)e^{f(Z)})}{e^{f(Z)}\pi(Z)+1-\pi(Z)}-\frac{\pi(Z)(1-R)(Y-\mu(Z))e^{f(Z)}}{e^{f(Z)}\pi(Z)+1-\pi(Z)}\right\}\right].$$
Then,
\begin{align*}
g_0(r)=& \bar{m}(\mu_0+rd_\mu, \pi_1+rd_\pi, f_0)\\
=&\E\left(Z\frac{(\pi_0(Z)-rd_\pi(Z))R(Y_1-(\mu_0(Z)+rd_\mu(Z))e^{f_0(Z)})}{e^{f_0(Z)}(\pi_1(Z)+rd_\pi(Z))+1-\pi_1(Z)-rd_\pi(Z)}\right)\\
&-\E\left(Z\frac{(\pi_1(Z)+rd_\pi(Z))(1-R)(Y_0-(\mu_0(Z)+rd_\mu(Z)))e^{f_0(Z)}}{e^{f_0(Z)}(\pi_1(Z)+rd_\pi(Z))+1-\pi_1(Z)-rd_\pi(Z)}\right)\\
=&\E\left(Z\frac{r^2e^{f_0(Z)}d_\pi(Z)d_\mu(Z)}{e^{f_0(Z)}\{\pi_1(Z)+rd_\pi(Z)\}+\pi_0(Z)-rd_\pi(Z)}\right),
\end{align*}
where   $f_0(z)=\delta_0^\top z.$ Let $\bar{\mu}(z)=\mu_0(z)+d_\mu(z)$ and $\bar{\pi}(z)=\pi_1(z)+d_\pi(z).$
Using a similar argument as above, because \eqref{eq:bounded-denom} is bounded for $\pi = \pi_1$ by Assumption~\ref{assume:problem-regularity}(c) and $\mu_1(z)$ is integrable by Assumption~\ref{assume:problem-regularity}(e), the Dominated Convergence Theorem yields
\begin{align*}
    \frac{dg_0(r)}{dr}=&2r\E\left(Z\frac{e^{f_0(Z)}d_\pi(Z)d_\mu(Z)}{e^{f_0(Z)}(\pi_1(Z)+rd_\pi(Z))+1-\pi_1(Z)-rd_\pi(Z)}\right)\\
    &-r^2\E\left(Z\frac{e^{f_0(Z)}(e^{f_0(Z)}-1)d_\pi(Z)^2d_\mu(Z)}{\{e^{f_0(Z)}(\pi_1(Z)+rd_\pi(Z))+1-\pi_1(Z)-rd_\pi(Z)\}^2}\right)
\end{align*}
Therefore
$$\frac{dg_0(r)}{dr}\biggm|_{r=0}=0,$$ which verifies Assumption~\ref{assume:dml-problem-reg}(d).

Assumption~\ref{assume:estimator-regularity} implies that there
exists sequences $\log(n)n^{-1/4} \le a_n=o(1),$  and $\Delta_n'=o(1),$ such that 
\begin{equation*}
  \| \what{\pi}(\cdot) - \pi_1(\cdot) \|_{P,2} + \| \what{\mu}(\cdot) - \mu_0(\cdot) \|_{P,2} \le a_n n^{-1/4}, 
\end{equation*}
with probability $1-\Delta_n'/2$. Note that $a_n$ can be chosen such that these hold when
$\what{\pi}$ and $\what{\mu}$ are estimated using only $(1- K^{-1})n$ (as opposed to
$n$) samples.  
Let 
\begin{align*}
\mathcal{T}_n =& \left\{(\pi, \mu) \mid \pi, \mu~\text{are measureable}, 
\pi(\cdot)\in [\epsilon_\pi, 1-\epsilon_\pi], \mu(\cdot)\in [\epsilon_\mu, \epsilon_\mu^{-1}], \right. \\
& \left.~\mbox{and}~
\| \pi(\cdot) - \pi_1(\cdot) \|_{P,2} +\| \mu(\cdot) - \mu_0(\cdot) \|_{P,2} \le a_n n^{-1/4} 
 \right\},
 \end{align*}
  Then, $P(\{(\what{\pi}^{(-k)}, \what{\mu}^{(-k)}) \in \mathcal{T}_n\}_{k=1}^K) \ge 1 - K\Delta_n'$. Let $\Delta_n = K\Delta_n'$, and Assumption~\ref{assume:dml-est-reg}(a) is satisfied.

For $Q \in \mathcal{P}$, and $(\mu, \pi) \in {\cal T},$
\vspace{-1em}
\begin{align*}
    &\| m(G; \widebar{\delta}, \pi, \mu)-m(G; \delta_0, \pi, \mu) \|_{Q,2} \\
    =& \left\| \widetilde{Z}\left(\pi(1-R)Y^{(0)} + \mu(R - \pi) \right) \left(\frac{\exp(\widebar{\delta}^\top \widetilde{Z})}{e^{\widebar{\delta}^\top \widetilde{Z}}\pi + (1-\pi)} - \frac{\exp(\delta_0^\top \widetilde{Z})}{e^{\delta_0^\top \widetilde{Z}}\pi + (1-\pi)} \right) \right\|_{Q,2}\\
    \le& \left\| \widetilde{Z}\left\{\pi\pi_0\mu_0 + \mu(\pi_1 - \pi) \right\} \right\|_{Q,2}\left\| (1-\pi)\frac{\exp(\widebar{\delta}_0^\top \widetilde{Z}) - \exp(\delta^\top \widetilde{Z})}{(e^{\widebar{\delta}^\top \widetilde{Z}}\pi + (1-\pi))(e^{\delta_0^\top \widetilde{Z}}\pi + (1-\pi))} \right\|_{Q,2}\\
    \le & C \left\| \exp(\widebar{\delta}_0^\top \widetilde{z}) - \exp(\delta^\top \widetilde{z}) \right\|_{\infty} \le  C L_{\text{rad}} \| \widebar{\delta} - \delta_0\|_{\infty},
\end{align*}
where we suppressed $Z$ in functions such as $\pi(Z)$, $\mu(Z)$, etc to simplify notation,  $L_{\text{rad}}$ is the Lipschitz constant of $t \mapsto \exp(t)$ over $|t| \le \sup_{\delta \in \Omega, z \in \mathcal{Z}} |\delta^\top \widetilde{z}|$. Therefore, $m(G; \delta, \mu, \pi)$ is Lipschitz in $\delta$. For all $(\mu, \pi) \in {\cal T},$ \eqref{eq:bounded-denom} and the fact that $\mu \in L_2(Q)$ imply that there exists a squared-integrable envelope function $F_{1,(\mu,\pi)}$. This, the Lipschitz constraint, and the bound $\log N(\epsilon_N, \Omega, \|\cdot\|_\infty) \le \widetilde{v} \log(\widetilde{a}/\epsilon_N)$ on the parameter space imply that $\sup_{Q} \log N(\epsilon_N\|F_{1,(\mu,\pi)}\|_{Q,2}, \mathcal{F}_{1,\mu,\pi}, \|\cdot\|_{Q,2}) \le v \log (a / \epsilon_N)$. Thus Assumption~\ref{assume:dml-est-reg}(b) is verified.
\begin{align*}
r_n=&\|\E \{m(G; \delta, \mu, \pi)\}-\E\{m(G; \delta, \mu_0, \pi_1)\}\|\\
=& \left\| \E\left[ \widetilde{Z}\frac{(\pi_1 - \pi)\exp(\delta^\top \widetilde{Z})(\pi_1(\mu_1 - \mu \exp(\delta^\top \widetilde{Z})) + (1 - \pi)(\mu_0 - \mu))}{\left( e^{\delta^\top \widetilde{Z}}\pi + (1 - \pi) \right)\left( e^{\delta^\top \widetilde{Z}}\pi_1 + (1 - \pi_1) \right)} \right] \right\| 
\le \widetilde{C}_1 \| \pi - \pi_1 \|_{P, 2},
\end{align*}
for a finite constant $\widetilde{C}_1.$ Therefore, by Assumption~\ref{assume:dml-est-reg}(c), we can choose $\tau_n = (a_n^{3/8} n^{-1/4})$ and $\delta_n=\sqrt{a_n}$ to satisfy $r_n\le \widetilde{C}_1\|\pi - \pi_1\|_{P,2} \le \delta_n \tau_n$ for adequately large $n,$ using the definition of $\mathcal{T}_n$. Next,
\begin{align*}
&\left\{\E \|m(G; \delta, \mu, \pi)-m(G; \delta_0, \mu_0, \pi_1)\|^2\right\}^{1/2}\\
\le&\sqrt{3}\left\{\E \|m(G; \delta, \mu, \pi)-m(G; \delta, \mu_0, \pi)\|^2\right\}^{1/2}
   +\sqrt{3}\left\{\E \|m(G; \delta, \mu_0, \pi)-m(G; \delta, \mu_0, \pi_1)\|^2\right\}^{1/2}\\
   &+\sqrt{3}\left\{\E \|m(G; \delta, \mu_0, \pi_1)-m(G; \delta_0, \mu_0,\pi_1)\|^2\right\}^{1/2}\\
=&\sqrt{3}\left( \E\left[ \|\widetilde{Z}\|^2\frac{\exp(2\delta^\top\widetilde{Z})\left\{\pi_1(1-\pi)^2+\pi_0\pi^2\right\}}{\left(  e^{\delta^\top \widetilde{Z}}\pi + (1 - \pi) \right)^2}(\mu-\mu_0)^2\right]\right)^{1/2}
\end{align*}
  \begin{align*}
&+\sqrt{3}\left( \E\left[ \|\widetilde{Z}\|^2\frac{\exp(2\delta^\top \widetilde{Z})\left\{\pi_1\left(Y^{(1)} - \mu_0 \exp(\delta^\top \widetilde{Z})\right)^2 + \pi_0\left(Y^{(0)} - \mu_0\right)^2\right\}}{\left( e^{\delta^\top \widetilde{Z}}\pi + (1 - \pi) \right)^2\left( e^{\delta^\top \widetilde{Z}}\pi_1 + \pi_0 \right)^2} (\pi_1-\pi)^2\right] \right)^{1/2}\\
&+\sqrt{3} \left(\E \left[ \|\widetilde{Z}\|^2\frac{\pi_1^2\pi_0^2\left\{\pi_1\left(Y^{(1)} - \mu_1\right)^2 + \pi_0\left(Y^{(0)} - \mu_0\right)^2+(\pi_1\mu_1+\pi_0\mu_0)^2/\pi_1 \right\}}{\left( e^{\delta^\top \widetilde{Z}}\pi + (1 - \pi) \right)^2\left( e^{\delta^\top \widetilde{Z}}\pi_1 + \pi_0 \right)^2} (e^{\delta^\top \widetilde{Z}}-e^{\delta_0^\top \widetilde{Z}})^2\right]\right)^{1/2}\\
\le & \widetilde{C}_2 \left(\|\pi-\pi_1\|_{P,2}+\|\mu-\mu_0\|_{P,2}+\|\delta-\delta_0\|_2\right)
\end{align*}
where  $(\mu, \pi) \in \mathcal{T}_n, \|\delta - \delta_0\|_2 \le \tau_n,$ $\widetilde{C}_2$ is a finite constant that depends on constants such as $\sigma_U^2$, $\epsilon_\pi$, $\epsilon_\mu$ in Assumption ~\ref{assume:problem-regularity}. 
Therefore, 
\begin{equation*}
    r_n' = \sup_{(\mu, \pi) \in \mathcal{T}_n, \|\delta - \delta_0\|_2 \le \tau_n}\left\{\E \|m(G; \delta, \mu, \pi)-m(G; \delta_0, \mu_0, \pi_1)\|^2\right\}^{1/2} \le \widetilde{C}_2 \left(a_n n^{-1/4} + \tau_n \right),
\end{equation*}
Thus, $r_n'\log^{1/2}(1/r_n')\le  \widetilde{C}_2 a_n^{3/8} n^{-1/4}\sqrt{\log(n)} \le \delta_n$ and, thus, Assumption~\ref{assume:dml-est-reg}(d) is satisfied. 

Let $d_f(z)=\widebar{f}(z)-f_0(z)=(\widebar{\delta}-\delta_0)^\top \widetilde{z}.$ Define
\begin{align*}
k(r)=& \bar{m}(\mu_0+rd_\mu, \pi_1+rd_\pi, f+rd_f)\\
=&\E\left(Z\frac{\left[(\pi_0-rd_\pi)R(Y_1-(\mu_0+rd_\mu)e^{f_0+rd_f})-(\pi_1+rd_\pi)(1-R)(Y_0-(\mu_0+rd_\mu))e^{f_0+rd_f}\right]}{e^{f_0+rd_f}(\pi_1+rd_\pi)+\pi_0-rd_\pi}\right)\\
=&\E \left( Ze^{f_0}\frac{\left[(1-e^{rd_f})\pi_0\pi_1\mu_0-r(1-e^{rd_f})\pi_1\mu_0d_\pi+r^2 e^{rd_f}d_\mu d_\pi\right]}{e^{f_0+rd_f}(\pi_1+rd_\pi)+\pi_0-rd_\pi} \right)\\
=&\E \left\{ Ze^{f_0(Z)}\frac{h_1(Z, r)}{h_2(Z, r)}\right\},
\end{align*}
where $h_1(z, r)=(1-e^{rd_f(z)})\pi_0(z)\pi_1(z)\mu_0(z)-r(1-e^{rd_f(z)})\pi_1(z)\mu_0(z)d_\pi(z)+r^2 e^{rd_f(z)}d_\mu(z) d_\pi(z),$
and 
$h_2(z, r)=e^{f_0(z)+rd_f(z)}\{\pi_1(z)+rd_\pi(z)\}+\pi_0(z)-rd_\pi(z).$ Similar as above, by the Dominated Convergence Theorem, we have
\begin{align*}
\frac{d^2 k(r)}{dr^2}=&\E \left( Z e^{f_0(Z)} \left[\frac{\partial^2 h_1(Z, r)/\partial r^2}{h_2(Z, r)}-2\frac{\partial h_1(Z, r)/\partial r \partial h_2(Z, r)/\partial r}{h^2_2(Z, r)} \right.\right. \\
&\left.\left.~~~~~~~~~~-\frac{h_1(Z, r) \partial^2 h_2(Z, r)/\partial r^2}{h^2_2(Z, r)}+2\frac{h_1(Z, r) \{\partial h_2(Z, r)/\partial r\}^2}{h^3_2(Z, r)}
\right]\right),
\end{align*}
where
\begin{align*}
\frac{\partial h_1(z, r)}{\partial r}=&-e^{rd_f}\pi_0\pi_1\mu_0d_f-(1-e^{rd_f})\pi_1\mu_0d_\pi+R_{11}(r, z; \kappa)\\
\frac{\partial^2 h_1(z, r)}{\partial r^2}=&-e^{rd_f}\pi_0\pi_1\mu_0d_f^2+R_{12}(r, z; \kappa)\\
\frac{\partial h_2(r)}{\partial r}=&(e^{f_0+rd_f}-1)d_\pi+e^{f_0+rd_f}\pi_1 d_f +R_{21}(r, z; \kappa)\\
\frac{\partial^2 h_1(r)}{\partial r^2}=&e^{f_0+rd_f}\pi_1d_f^2+R_{22}(r, z; \kappa),
\end{align*}
$\kappa=(d_\pi, d_\mu, d_f)^\top,$ and $R_{ij}(r, z; \kappa)$ is a function of $r, z$ satisfying that 
$$\sup_{(r,z)\in [0, 1]\times{\cal Z}} \frac{|R_{ij}(r, z; \kappa)|}{|d_\pi(z)d_\mu(z)|+|d_f(z)d_\pi(z)|}\le \widetilde{C}_3 $$ 
for a  constant $\widetilde{C}_3.$  Therefore, after careful regrouping,
\begin{align*}
\lambda_n=&\sup_{r\in (0,1), (\bar{\mu}, \bar{\pi})\in {\cal T}_n, |\bar{\delta}-\delta_0|\le \tau_n}\left\|\frac{d^2 k(r)}{dr^2}\right\|\\
\le & \widetilde{C}_4 (\|d_f\|^2_{P,2}+\|d_f\|_{P,2}\|d_\pi\|_{P,2}+\|d_\pi\|_{P,2} \|d_\mu\|_{P,2}+\|d_\pi\|^2_{P,2}),\\
\le & \tilde{C}_5 (\tau_n^2+\tau_n a_n n^{-1/4}+a_n^2n^{-1/2}) \le \sqrt{a_n} n^{-1/2}=\delta_nn^{-1/2}
\end{align*}
where $\widetilde{C}_i$ are finite constants. Thus,  Assumption~\ref{assume:dml-est-reg}(d) is verified. 

Lastly, to verify Assumption~\ref{assume:dml-est-reg}(e), note that
\begin{align*}
&\E\left[m(G; \delta_0, \mu_0, \pi_1)m^\top (G; \delta_0, \mu_0, \pi_1)\right]\\
=& \E\left\{\widetilde{Z}\widetilde{Z}^\top \pi_0(Z)\pi_1(Z) \frac{\pi_0(Z)\mbox{var}(Y^{(1)} \mid Z)+\pi_1(Z)\mbox{var}(Y^{(0)}\mid Z)e^{2\delta_0^\top\widetilde{Z}}}{\left\{e^{\delta_0^\top\widetilde{Z}}\pi_1(Z)+\pi_0(Z)\right\}^2}\right\},
\end{align*} 
which is non-degenerate, because Assumption~\ref{assume:problem-regularity} ensures that 
$\mbox{var}(Y^{(r)}\mid Z=z)\ge \sigma_L>0.$ 

Note that our choice of $\delta_n$ and $\tau_n$ satisfies $\log(n)/\sqrt{n} \le a_n^{3/8}n^{-1/4}=\tau_n \le \sqrt{a_n}=\delta_n$, and $\delta_n=\sqrt{a_n} \ge \log(n)^{-1/2}\ge n^{-1/2 + 1/q}\log(n)$ for any constant $q>2.$ Therefore, all assumptions of Theorem~\ref{thm:chernozhukov} are verified. Applying this theorem completes the proof.

\end{proof}

\newpage
\begin{table}[!htbp]
    \caption{Baseline characteristics of RRMS patients at the initiation of therapy with DMF and TERI: mean (standard deviation) for continuous covariate and number (proportion) for binary covariate}
    \label{table:baseline}
      \centering
      \begin{tabular}{cccc}
      \hline \hline
      variable &  TERI $(n=1050)$  & DMF $(n=1741)$  & $p$ value\\
      \hline
exposure time  (year)       & 2.11(1.71)   & 2.17(1.72)  & 0.603\\
age                         & 44.86(10.20) & 39.91(10.74)& 0.0000\\
\# prior treatments         & 0.97(0.93)   & 0.96(0.98)  & 0.4703\\
MS duration (year)          & 8.11(7.64)   & 6.57(6.60)  & 0.0000\\
GA                          &  821(78.2\%) & 1327(76.2\%)     & 0.246\\
IFN                         &  502(47.8\%) & 886(50.9\%)      & 0.118\\
\# relapses (prior year)    & 0.42(0.60)   &  0.46(0.65) & 0.2032\\
\# relapses (prior 2 years) & 0.64(0.84)   &  0.71(0.90) & 0.095\\
EDSS                        & 2.03(1.51)   & 1.84(1.50)  & 0.0006\\
pyramidal EDSS              & 0.92(1.10)   & 0.77(1.04)  & 0.0000\\
\hline
\end{tabular}
\end{table}

\newpage
\begin{table}[!htbp]
    \caption{The estimated weights in constructed CATE scores (TERI vs DMF) }
    \label{table:score}
      \centering
      \begin{tabular}{c rrr}
  \hline\hline
 & \multicolumn{3}{c}{ratio of relapse rate} \\
  \cline{2-4} 
 & na\"ive reg. & two reg. &  contrast reg. \\
 \hline
intercept                     & 0.692 & 0.476 & 0.670 $(0.711)^*$\\
age                           & 0.013 & 0.013 & 0.017 (0.013)$~$ \\
\# prior treatments           &-0.303 & 0.011 &-0.088 (0.195)$~$ \\
MS duration                   & 0.022 & 0.045 & 0.028 (0.028)$~$ \\
(years)                       &&&\\       
GA                            &-0.584 & 0.517 &-0.700 (0.349)$~$ \\
IFN                           &-0.304 &-0.024 &-0.185 (0.318)$~$ \\
\# relapses                   &-0.258 &-0.661 &-0.811 (0.271)$~$ \\
(prior year)                  &&&\\
\# relapses                   & 0.191 & 0.360 & 0.444 (0.201)$~$ \\
(prior two years)             &&&\\
EDSS                          &-0.046 &-0.247 &-0.233 (0.114)$~$ \\
pyramidal EDSS                & 0.006 & 0.027 &-0.004 (0.160)$~$ \\
\hline
\multicolumn{4}{l}{$^*:$ the estimated standard error of the weight.}
\end{tabular}
\end{table}

\newpage
\begin{table}[!htbp]
    \caption{Empirical bias and coverage level of the 95\% confidence interval in estimating $\delta_0$ based on 400 replications under the well-specified contrast and well-specified Poisson simulation settings: }
    \label{table:simulation}
      \centering
      \begin{tabular}{l c rr c c rr} 
      \hline \hline
      Covariates & \multicolumn{3}{c}{Well-specified Poisson} && \multicolumn{3}{c}{Well-specified Contrast} \\
       & Coefficients      &  Bias & Coverage &&Coefficients & Bias & Coverage \\
               \cline{2-4} \cline{6-8}\\ 
Intercept &0.375      &  0.005&93.0\% &&-0.10& 0.000&92.5\%\\                       
$Z_1$ &0.125      & -0.021&93.8\% &&0.25 &-0.013&91.3\%\\
$Z_2$ &0.050      & -0.009&95.0\% &&0    &-0.007&92.0\%\\
$Z_3$ &0          & -0.003&94.5\% &&0    &-0.007&95.0\%\\
$Z_4$ &0          &  0.003&93.0\% &&0    & 0.003&94.5\%\\
$Z_5$ &0          &  0.005&91.8\% &&0    & 0.003&93.5\%\\
$Z_6$ &-0.25      &  0.004&93.8\% &&0.25 &-0.009&94.0\%\\
$Z_7$ &0          &  0.005&92.5\% &&0    & 0.006&95.0\%\\
$Z_8$ &0          &  0.000&93.5\% &&0    &-0.001&92.8\%\\
$Z_9$ &0          & -0.001&94.0\% &&0    &-0.002&95.3\%\\
$Z_{10}$ &0       & -0.004&93.0\% &&0    &-0.003&93.5\%\\
\hline
\end{tabular}
\end{table}

\newpage
\begin{figure}[htpb!]
\centering
\caption{The ATE in subgroups of patients identified by different CATE scores in four simulation settings including: the true CATE, contrast regression, two regression, na\"ive regression, boosting, modified outcome boosting, and Bayesian additive regression trees (BART).}
  \centering
  \includegraphics[scale = 0.7]{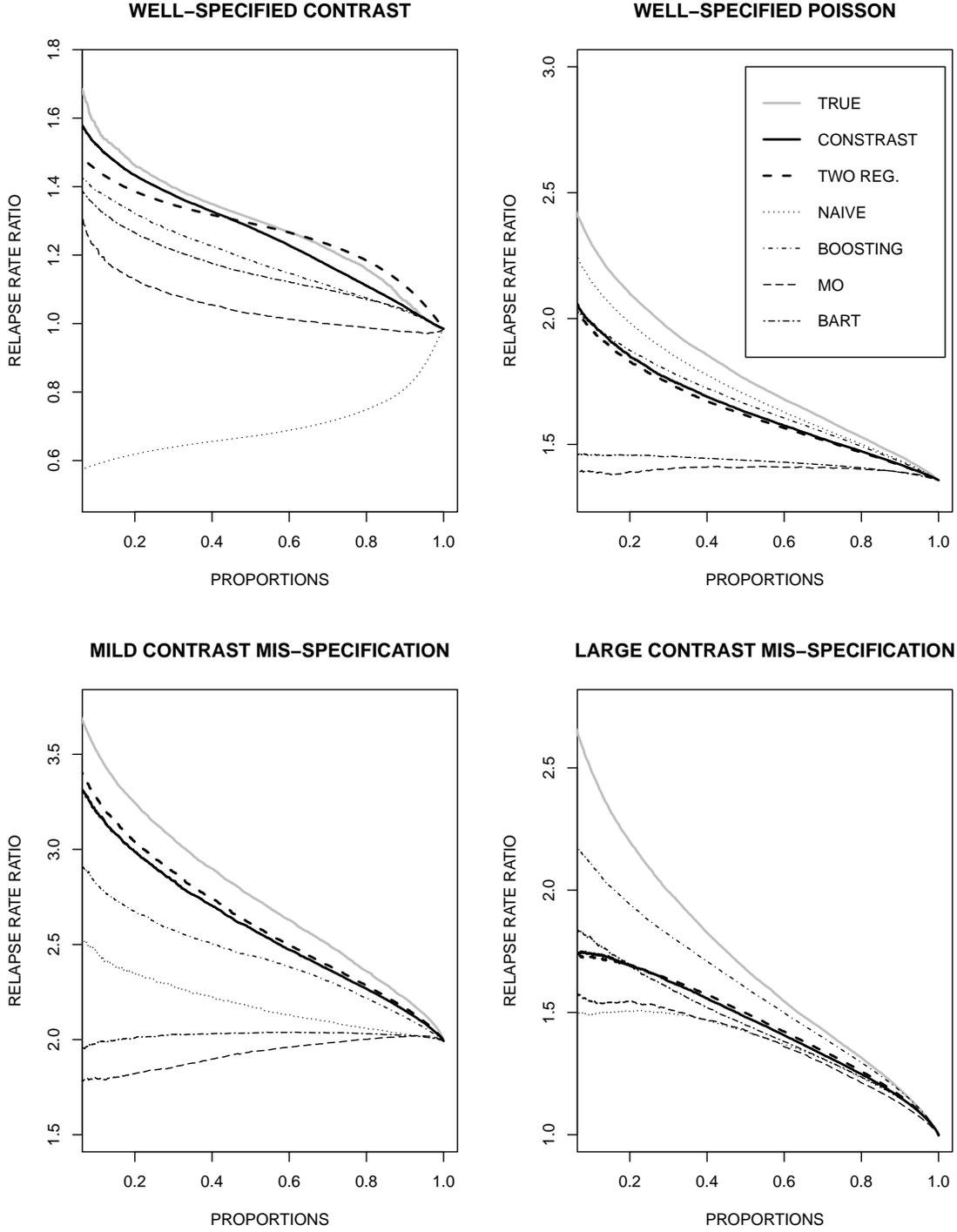}
  \label{fig:simVC}
\end{figure}

\newpage
\begin{figure}[htpb!]
\centering
\caption{The distribution of correlation coefficients between the estimated and true CATE in four simulation settings; there are six methods considered from left to right: contrast regression (dark gray), two regression (light gray), na\"ive regression, boosting, modified outcome boosting, and Bayesian additive regression trees (BART).}
  \centering
  \includegraphics[scale = 0.7]{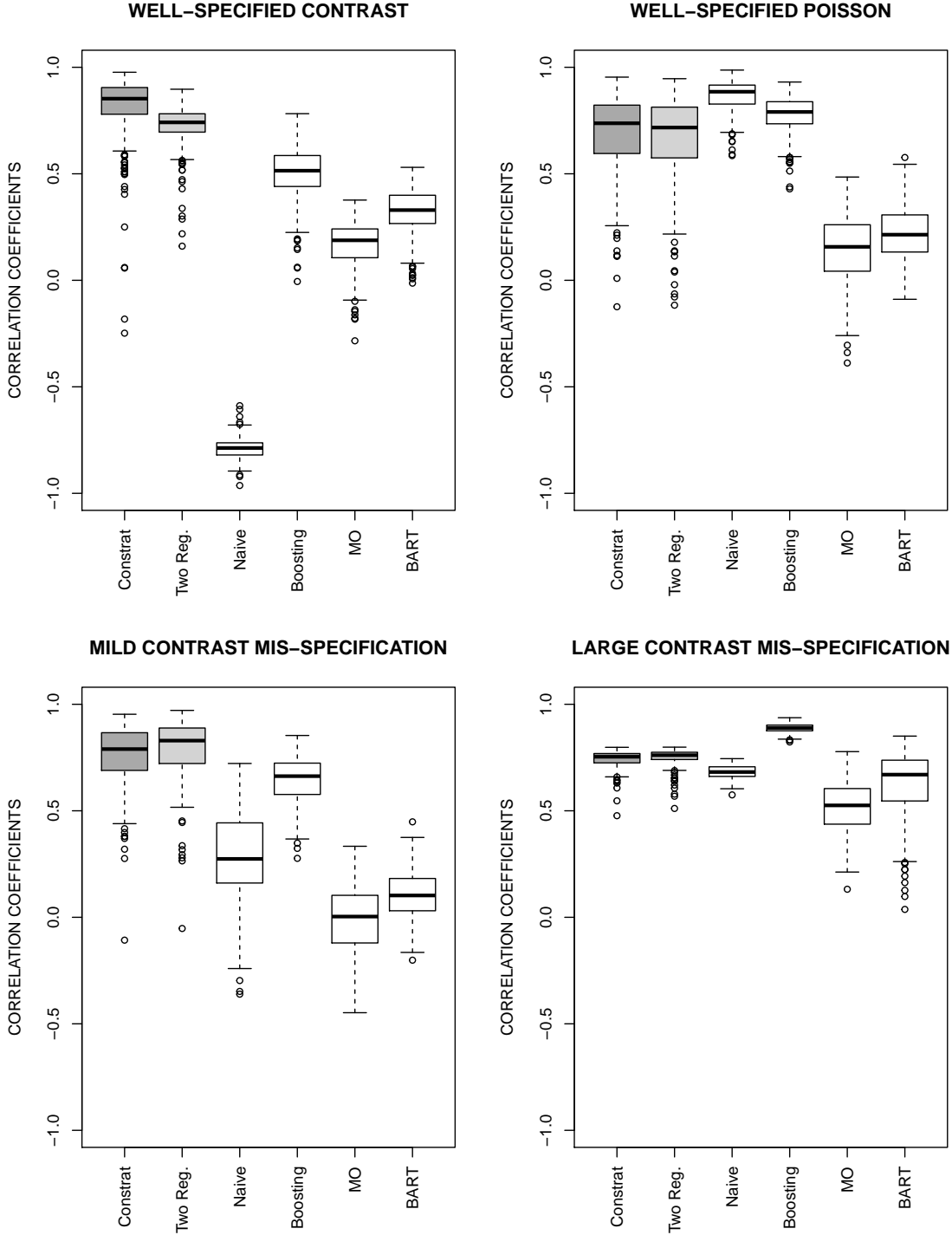}
  \label{fig:simCC}
\end{figure}

\newpage
\begin{figure}[htpb!]
\centering
\caption{The log-transformed CATE scores based on the standard regression approach and the proposed doubly robust adjustment method: the CATE for the ratio of relapse rates}
  \centering
  \includegraphics[scale =0.6]  {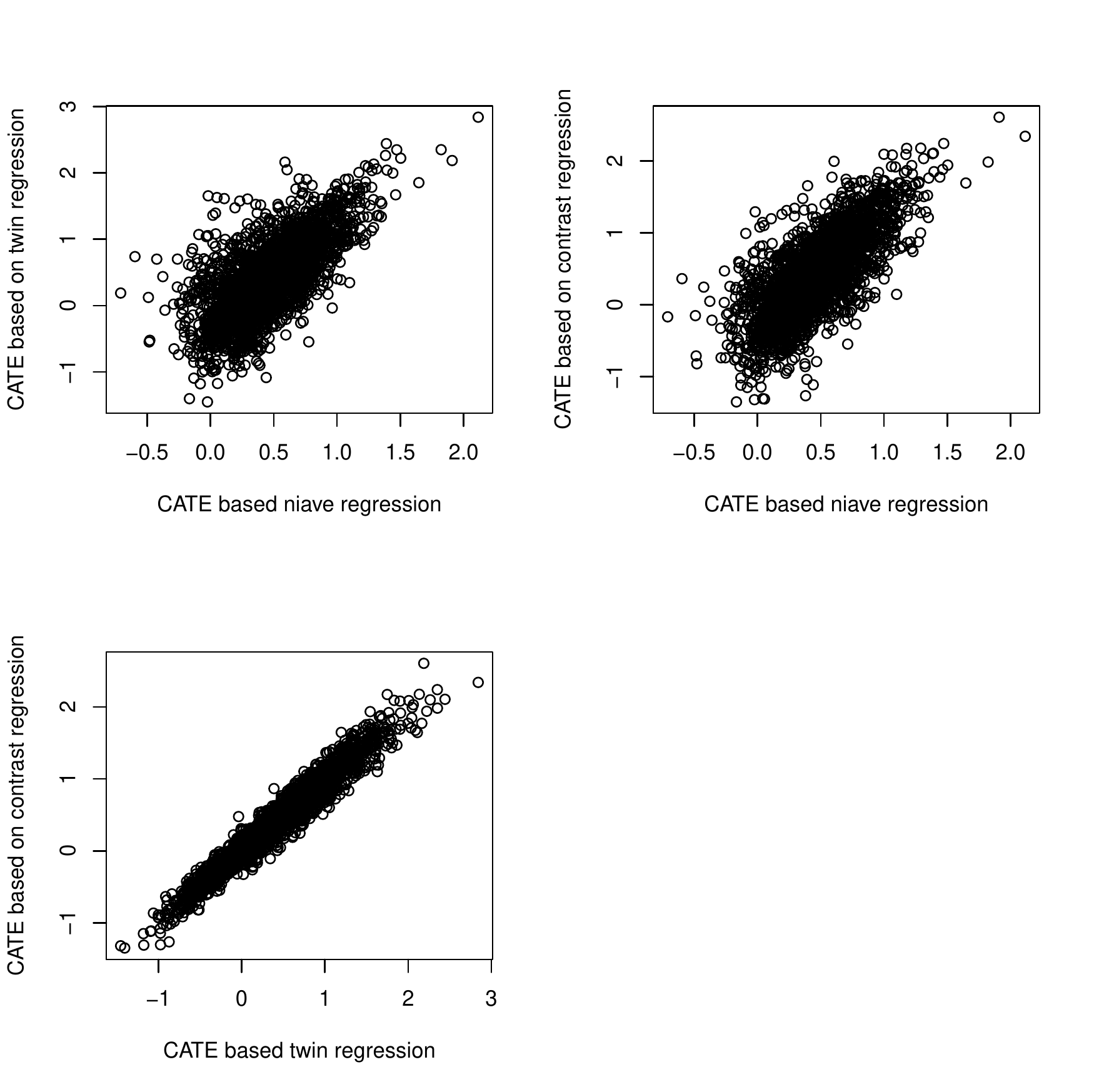}
  \label{fig:cor}
\end{figure}

\newpage
\begin{figure}[htpb!]
\centering
\caption{The ATE (relpase rate ratio of TERI vs DMF) in subgroups of patients based on the CATE scores constructed in the training set (two proposed methods, na\"ive regression and boosting) in the NTD registry}
  \centering
  \includegraphics[scale =0.6]{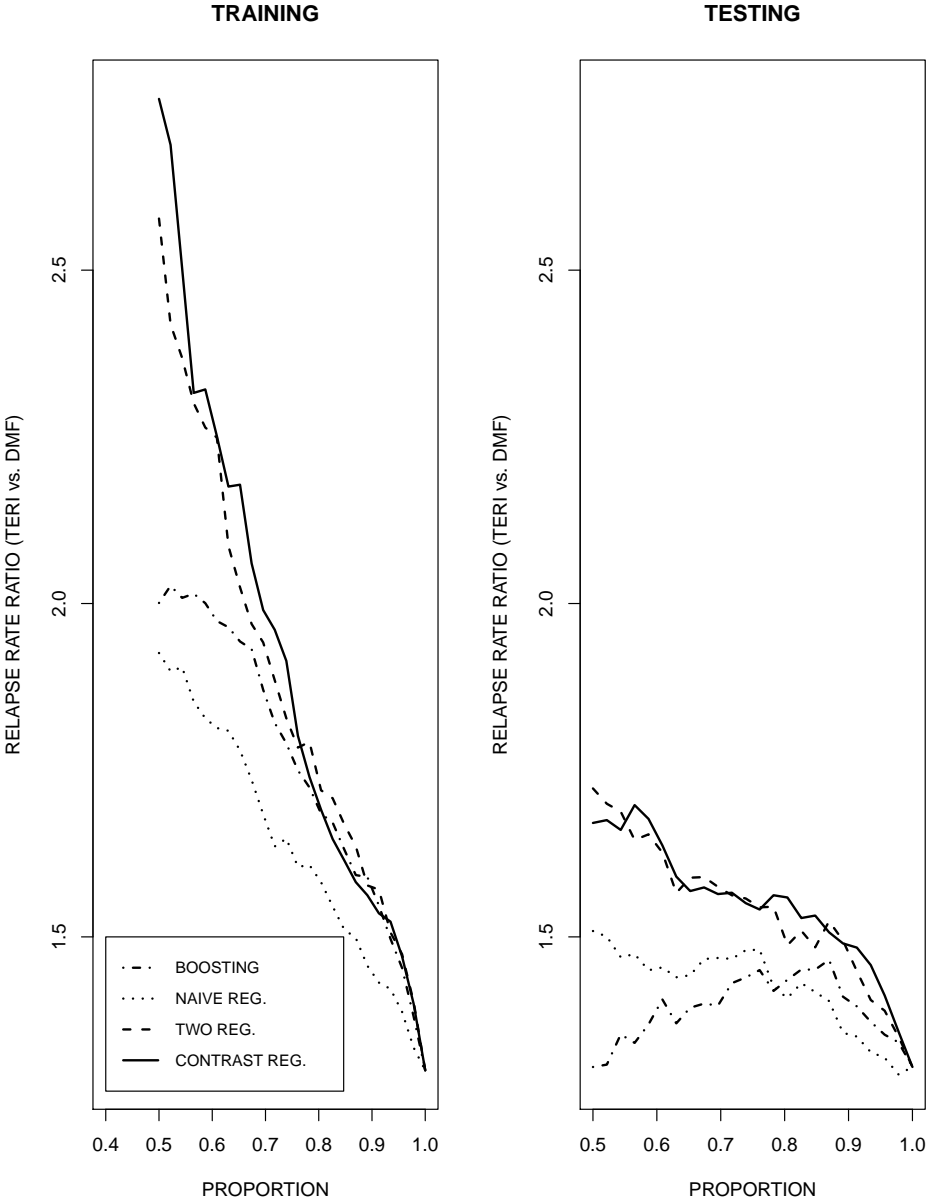}  
  \label{fig:exp1}
\end{figure}

\newpage
\begin{figure}[htpb!]
\centering
\caption{The cross-validated ATE (relapse rate ratio of TERI vs DMF) of subgroups of patients identified by different CATE scores (two proposed methods, na\"ive regression and boosting ) in the NTD registry.}
  \centering
  \includegraphics[scale=0.6]{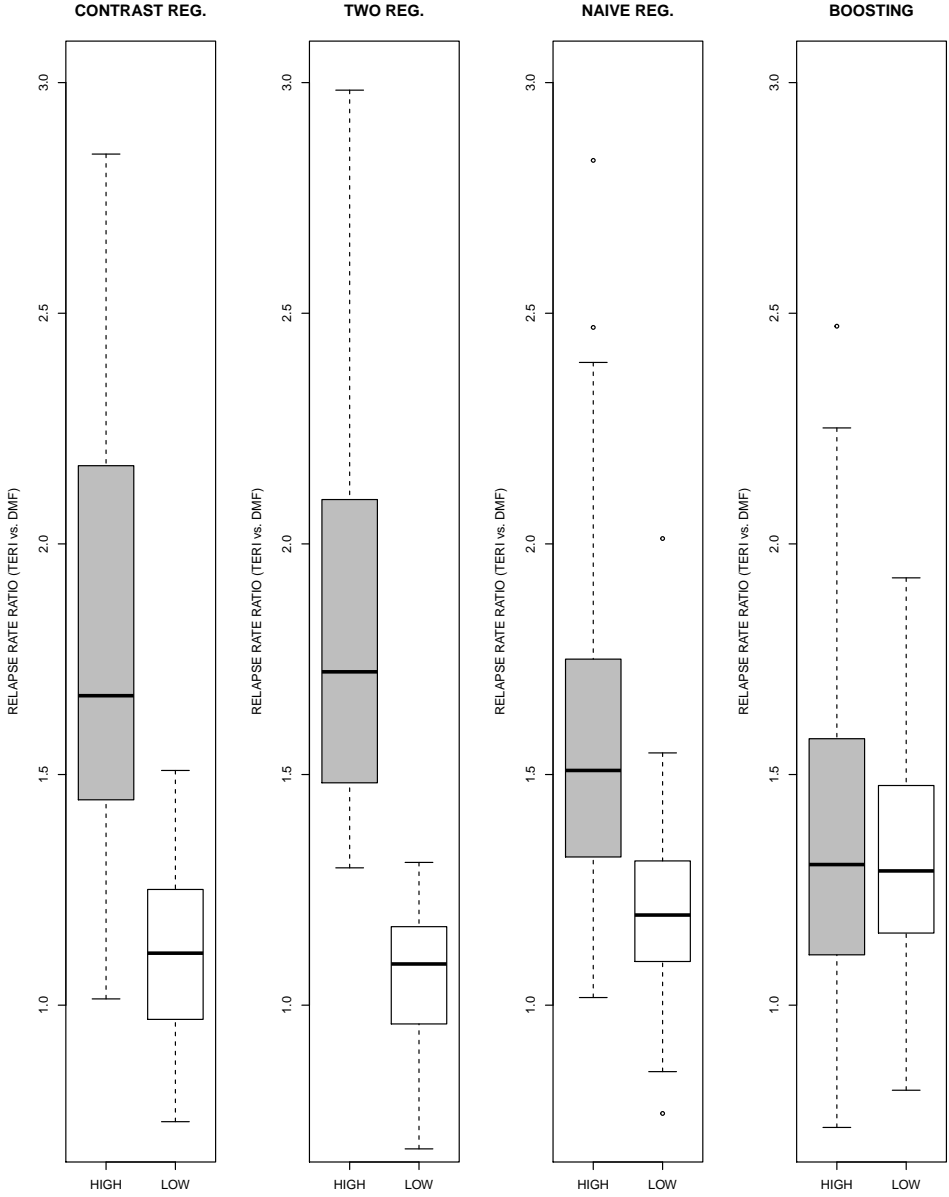}
  \label{fig:exp2}
\end{figure}

\renewcommand{\thesection}{\arabic{section}}
\setcounter{section}{0}

\section*{Supplementary Materials}

\section{The Optimal Weights for the Estimating Equation}
Consider all weighted estimating equations in the form of
\begin{align}
&S_n(\delta; w, \mu) \nonumber \\
=&n^{-1}\sum_{i=1}^n \widetilde{Z}_i w(Z_i)\left(\left[\{1-\pi(Z_i)\}R_iY_i^{(1)}-\pi(Z_i)(1-R_i)Y_i^{(0)}e^{\delta^\top \widetilde{Z}_i}\right]-\mu(Z_i)\{R_i-\pi(Z_i)\}\right). \label{eq:est-class}
\end{align}
If $\pi(z)=\pi_1(z)=P(R=1\mid Z=z)$ the correct propensity score, the estimating function converges to 
$$s(\delta; w, \mu)=\E\left[w(Z)\pi_1(Z)\pi_0(Z)\{\mu_1(Z)-\mu_0(Z)\exp(\delta_0^\top Z)\}\right]$$
and $\delta_0$ is the root of $s(\delta; w, \mu)=0.$  Thus, the root of the estimating equation $S_n(\delta; w, \mu)=0,$ denoted by $\what{\delta}_{w,\mu},$ is consistent in estimating $\delta_0,$ and  
$\sqrt{n}(\what{\delta}_{w,\mu}-\delta_0)$ converges to a mean zero Gaussian distribution with a variance-covariance matrix of  $A_{w,\mu}^{-1}B_{w,\mu}A_{w,\mu}^{-1},$
where
$$A_{w,\mu}=\E\left\{\tilde{Z}\tilde{Z}^{\top} w(Z)\pi_1(Z)\pi_0(Z)\mu_1(Z)  \right\},$$
and
$$B_{w, \mu}=\E\left\{\tilde{Z}\tilde{Z}^{\top} w(Z)^2 \left(\left[\pi_0(Z_i)R_iY_i^{(1)}-\pi_1(Z_i)(1-R_i)Y_i^{(0)}e^{\delta_0^\top \widetilde{Z}_i}\right]-\mu(Z_i)\{R_i-\pi_1(Z_i)\}\right)^2 \right\}.$$
First, since 
$$\E\left(\left[\pi_0(Z_i)R_iY_i^{(1)}-\pi(Z_i)(1-R_i)Y_i^{(0)}e^{\delta_0^\top \widetilde{Z}_i}\right]\mid Z_i=z\right)=0$$
and
$$ \E\left(R_i-\pi_1(Z_i)\mid Z_i=z\right)=0,$$
$B_{w, \mu}-B_{w, \bar{\mu}}$ is semi-positive definite, where
\begin{align*}
\bar{\mu}(z)=&\frac{\E\left(\left[\pi_0(Z_i)R_iY_i^{(1)}-\pi_1(Z_i)(1-R_i)Y_i^{(0)}e^{\delta_0^\top \widetilde{Z}_i}\right]\left\{R_i-\pi_1(Z_i)\right\} \mid Z_i=z \right)}{\E\left(\left\{R_i-\pi_1(Z_i)\right\}^2\mid Z_i=z\right)}\\
=& \frac{\pi_1(z)\pi_0(z)\mu_1(z)}{\pi_1(z)\pi_0(z)}=\mu_1(z).
\end{align*}
Secondly, by Cauchy's ineqaulity,
$A_{w,\mu_1}^{-1}B_{w,\mu_1}A_{w,\mu_1}^{-1}-A_{w_0,\mu_1}^{-1}B_{w_0,\mu_1}A_{w_0,\mu_1}^{-1}$ is also semi-positive definite, where
\begin{align*}
w_0(z)=& \frac{\E\left(\pi_1(Z)\pi_0(Z)\mu_1(Z)\mid Z=z\right)}{\E\left(\left[\pi_0(Z)RY^{(1)}-\pi_1(Z)(1-R)Y^{(0)}e^{\delta_0^\top \widetilde{Z}}-\mu_1(Z)\{R-\pi_1(Z)\}\right]^2 \mid Z=z\right)}\\
=& \frac{\pi_1(z)\pi_0(z)\mu_1(z)}{\E\left(\left[\pi_0(z)R(Y^{(1)}-\mu_1(z)\right]^2+\left[\pi_1(z)(1-R)\left\{Y^{(0)}-\mu_0(z)\right\}e^{\delta_0^\top \widetilde{z}}\right]^2 \mid Z=z\right)}\\
=& \frac{\pi_1(z)\pi_0(z)\mu_1(z)}{\pi_0(z)^2\pi_1(z)\mbox{var}(Y^{(1)}\mid Z=z)+\pi_0(z)\pi_1(z)^2\mbox{var}(Y^{(0)}\mid Z=z)e^{2\delta_0^\top z}}\\
=& \frac{\mu_1(z)}{\pi_0(z)\mu_1(z)+\pi_1(z)\mu_0(z)e^{2\delta_0^\top z}}=\frac{1}{\pi_0(z)+\pi_1(z)e^{\delta_0^\top z}},
\end{align*}
where we used the fact that $\mbox{var}(Y^{(r)}\mid Z=z)=\mu_r(z).$
Therefore
$A_{w,\mu}^{-1}B_{w,\mu}A_{w,\mu}^{-1}-A_{w_0,\mu_1}^{-1}B_{w_0,\mu_1}A_{w_0,\mu_1}^{-1}$ is always semi-positive definite, suggesting that an estimating function $S_n(\delta, w_0, \mu_1)$  with a limit
$$\E\left[\widetilde{Z} w_1(\delta; Z, \pi_1)\left\{\mu_1(Z)-\mu_0(Z)e^{\delta^\top \widetilde{Z}}\right\}\right],$$
yields the optimal variance among estimators of $\delta_0$ of the form \eqref{eq:est-class}.

\section{Average of symmetric contrast regression estimators}
The symmetric estimating equation for the one presented in Section 2.2.1 of the main paper, where we first estimate $\mu_1$ and $\pi_0$ was mentioned in Remark 2 of the main paper.
Averaging the original estimating equation with its counterpart based on $\mu_0$ and $\pi_1$ yields two new equations:
\begin{align*}
    \bar{S}_n(\delta)=&n^{-1}\sum_{i=1}^n \widetilde{Z}_i R_i\left[Y_i-\frac{e^{\delta^\top \widetilde{Z}_i}\widehat{\mu}_0(Z_i)+\widehat{\mu}_1(Z_i)}{2}\right]\frac{\what{\pi}_0(Z_i)}{e^{\delta^\top \widetilde{Z}_i}\what{\pi}_1(Z_i)+\what{\pi}_0(Z_i)}\\
    &-\sum_{i=1}^n \widetilde{Z}_i(1-R_i)\left[Y_ie^{\delta^\top \widetilde{Z}_i}-\frac{e^{\delta^\top \widetilde{Z}_i}\widehat{\mu}_0(Z_i)+\widehat{\mu}_1(Z_i)}{2}\right]\frac{\what{\pi}_1(Z_i)}{e^{\delta^\top \widetilde{Z}_i}\what{\pi}_1(Z_i)+\what{\pi}_0(Z_i)}=0,
\end{align*}
and
\begin{align*}
    \bar{S}^{(cf)}_n(\delta)=&n^{-1}\sum_{k=1}^K\sum_{i\in {\cal I}_k} \left(\widetilde{Z}_i R_i \frac{\left[Y_i-\{e^{\delta^\top \widetilde{Z}_i}\widehat{\mu}^{(-k)}_0(Z_i)+\widehat{\mu}^{(-k)}_1(Z_i)\}/2\right]}{e^{\delta^\top \widetilde{Z}_i}\widehat{\pi}_1^{(-k)}(Z_i)+\widehat{\pi}_0^{(-k)}(Z_i)}\widehat{\pi}_0^{(-k)}(Z_i) \right.\\
    & \left.-\widetilde{Z}_i(1-R_i)\frac{\left[Y_ie^{\delta^\top \widetilde{Z}_i}-\{e^{\delta^\top \widetilde{Z}_i}\widehat{\mu}^{(-k)}_0(Z_i)+\widehat{\mu}^{(-k)}_1(Z_i)\}/2\right]}{e^{\delta^\top \widetilde{Z}_i}\widehat{\pi}_1^{(-k)}(Z_i)+\widehat{\pi}_0^{(-k)}(Z_i)}\widehat{\pi}_1^{(-k)}(Z_i)\right)=0.
\end{align*}
In practice, we recommend to solve $\bar{S}^{(cf)}_n(\delta)=0$ for estimating $\delta_0$ due to the symmetry between $\mu_1$ and $\mu_0.$

\section{Proof of Theorem 2} 
First, for $c > 0$, we show that $AD(c) \ge c$, by showing that $\E(Y^{(1)} \mid D(Z) \ge c) \ge c \E( Y^{(0)} \mid D(Z) \ge c)$.
This follows from the following application of the tower property of conditional expectations,
\begin{align*}
    \E(Y^{(1)} \mid D(Z) \ge c) &= \E\{ \E(Y^{(1)} \mid Z) \mid D(Z) \ge c\} \\
    &= \E\{ D(Z) \E(Y^{(0)} \mid Z) \mid D(Z) \ge c\} \\
    &\ge \E\{ c \E(Y^{(0)} \mid Z) \mid D(Z) \ge c\} \\
    &= c \E\{ \E(Y^{(0)} \mid Z) \mid D(Z) \ge c\} \\
    &= c \E\{ Y^{(0)} \mid D(Z) \ge c\}.
\end{align*}

Now, let $0 < c' < c$. Expanding $AD(c')$ gives
\begin{align*}
    AD(c') &= \frac{\E\{ Y^{(1)} \mid D(Z) \ge c'\}}{\E\{ Y^{(0)} \mid D(Z) \ge c'\}} \\
    &= \frac{\E\{ Y^{(1)} \mid D(Z) \ge c\}p + \E\{ Y^{(1)} \mid D(Z) \in [c', c)\}(1-p)}{\E[ Y^{(0)} \mid D(Z) \ge c\}p + \E\{ Y^{(0)} \mid D(Z) \in [c', c)\}(1-p)},
\end{align*}
where $p = P\{D(Z) \ge c \mid D(Z) \ge c'\}.$ Now, consider the expression in the numerator, which we expand and bound as
\begin{align*}
    &\E\{ Y^{(1)} \mid D(Z) \ge c\}p + \E\{ Y^{(1)} \mid D(Z) \in [c', c)\}(1-p) \\
    &~~~~~~~= AD(c)\E\{ Y^{(0)} \mid D(Z) \ge c\}p + \E\{ Y^{(1)} \mid D(Z) \in [c', c)\}(1-p)\\
    &~~~~~~~= AD(c)\E\{ Y^{(0)} \mid D(Z) \ge c\}p  + \E\{ D(Z) \E(Y^{(0)} \mid Z) \mid D(Z) \in [c', c)\}(1-p)\\
    &~~~~~~~\le AD(c)\E\{ Y^{(0)} \mid D(Z) \ge c\}p  + c\E\{ Y^{(0)} \mid D(Z) \in [c', c)\}(1-p)\\
    &~~~~~~~\le AD(c)\left[\E\{ Y^{(0)} \mid D(Z) \ge c\}p  + \E\{ Y^{(0)} \mid D(Z) \in [c', c)\}(1-p)\right],
\end{align*}
where in the last inequality we used the fact that $AD(c) \ge c$. Plugging this into the expression for $AD(c')$ gives
\begin{equation*}
    AD(c') \le \frac{ AD(c) (\E\{ Y^{(0)} \mid D(Z) \ge c\}p  + \E\{ Y^{(0)} \mid D(Z) \in [c', c)\}(1-p))}{\E\{ Y^{(0)} \mid D(Z) \ge c\}p  + \E\{ Y^{(0)} \mid D(Z) \in [c', c)\}(1-p)} = AD(c).
\end{equation*}

\section{Example of CATE vs ATE Measured by OR}

\begin{figure}[htpb!]
\centering
\caption{The marginal ORs in nested subgroups according to the size of conditional OR (thick solid curve) and marginal ORs in nested subgroups maximizing the marginal ORs consecutively (thick dotted curve). The conditional OR (thin solid curve) is also plotted.}
  \centering
  \includegraphics[scale = 0.7]  {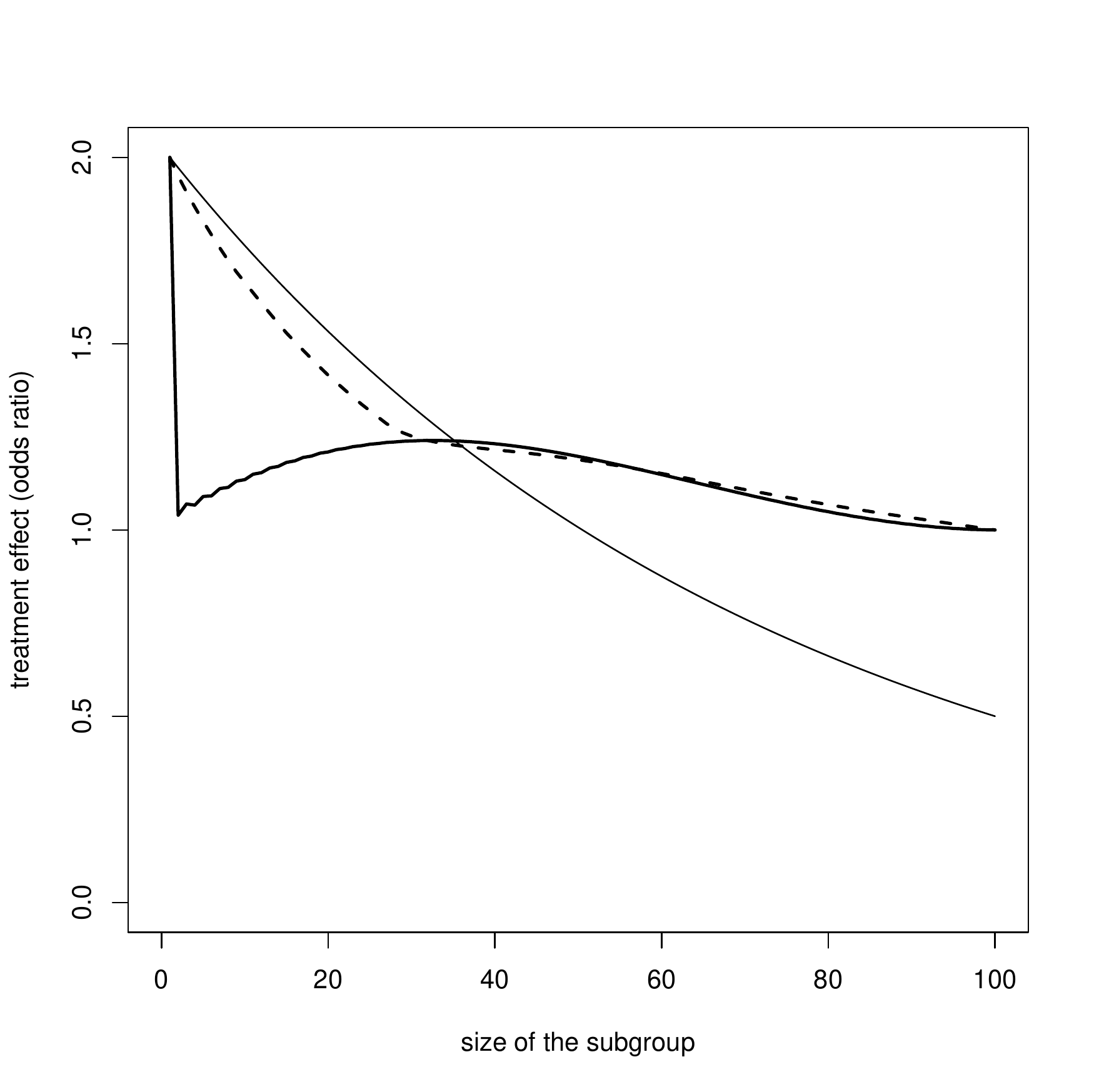}
  \label{fig:exmp2}
\end{figure}

Consider the case where the response rate of a binary outcome in the control arm is 
$$
p_i=\begin{cases} 1-\frac{i}{n+1}  ~\mbox{ if } ~ i ~\mbox{is even}, \\ \frac{i}{n+1} ~\mbox{ if } ~ i ~\mbox{is odd}, \end{cases}
$$
the response rate in the treatment arm is $p_i\theta_i/(1-p_i+p_i\theta_i),$  $\log(\theta_i)=\log(2)-\log(4)(i-1)/(n-1),$ and $n=100.$  Thus, these $100$ patients are sorted according to  their conditional OR, which is monotone decreasing from 2 to -2.  The marginal OR in 10 patients with the highest conditional OR is only 1.14, though the conditional OR of each patient is above $ 1.76.$ On the other hand,  we can select a different subgroup of 10 patients, in which the marginal OR is as high as 1.66. Figure \ref{fig:exmp2} plots the marginal ORs in a sequence of subgroups based on (1) the optimal grouping maximizing the marginal ORs and (2) the grouping according to the conditional OR of each patient. The former can be substantially higher than the latter.

\section{Time to Event Outcomes}
In precision medicine, many clinically meaningful outcomes to are time-to-event outcomes. For example, among MS patients in the NTD registry, a useful endpoint to measure effectiveness is the time to the first relapse. In this case, the observed training data consist of $\{(X_i, \Delta_i, R_i, Z_i), i=1,\cdots, n\},$ where $X_i=T_i\wedge C_i,$ $\Delta_i=I(T_i<C_i),$ $T_i$ is the event time of interest and $C_i$ is the censoring time. As in the main paper, we will use $T_i^{(r)}$ and $C^{(r)}_i$ to refer to the potential event time and censoring time under treatment $r$. We assume non-informative censoring, in the sense that $T_i^{(r)}$ and $C_i^{(r)}$ are independent, conditional on $Z_i$. For time-to-event outcomes, the heterogeneous treatment effect often cannot be measured by a contrast of the conditional expectations $\E(T^{(r)}|Z=z),$ because this expectation may not be identifiable due to right censoring preventing observation of the event time beyond the maximum study followup time. The parameter closest to the mean survival time that is identifiable is the restricted mean survival time, i.e., $E(T\wedge \tau)$, for an appropriate choice of $\tau > 0$. The ATE (and CATE) of the restricted mean survival time is an appropriate contrast that we can estimate with data. 

It is a popular practice to employ Cox proportional hazards model to analyze time-to-event outcomes. Let $\lambda_r(t \mid z) $ 
be the hazard function of $T_i^{(r)} \mid Z_i=z$, $r=0, 1.$ The Cox proportional hazards model assumes that $$\lambda_r(t \mid z)=\lambda_{r0}(t)\exp(\beta^\top_rz),$$
for $\lambda_{r0}(\cdot)$, an unspecified baseline hazard function in group $r$. In this case, $\exp\{(\beta_1-\beta_0)^\top z\}$ is a natural measure of the CATE. If the proportional hazards assumption is violated, we can still define $\beta_r$ as the root of the limit of the score equation of the Cox model.
However, in this case, $\exp\{(\beta_1-\beta_0)^\top z\}$ may no longer be a sensible surrogate for the CATE even if the two arms have the same covariate distribution, as it may introduce spurious heterogeneity not present in real data. 
We provide an example in Supplementary Section \ref{section:exampleHR}.

In this section, we propose to use the ratio of restricted mean time lost (RMTL) due to the first relapse to measure the treatment effect, i.e.,
$$D(z)=\frac{\E\{\tau-(T^{(1)}\wedge \tau)|Z=z\}}{\E\{\tau-(T^{(0)}\wedge \tau) \mid Z=z\}},$$
where $\tau>0$ is a chosen constant \citep{uno2014moving}, since it is expected to be concordant with the relapse rate ratio. When the relapse rate is low, this ratio is similar to the hazard ratio.

Again, we propose two approaches for approximating $D(z)$. The first is based on the semiparametric model for the contrast of the two arms, while the second is based on performing two regressions, adjusted so that they mimic the arms of an RCT without censoring. 

In the first approach, we may directly estimate $\delta_0$ under the assumption that $D(z)=\exp(\delta_0^\top \widetilde{z}).$ Specifically, $\delta_0$ is the solution to the estimating equation 
\begin{align*}
    & \E\left[ \frac{\mathbf{1}\{T \wedge \tau < C\}}{K_{C^{(R)}}(T \wedge \tau \mid Z)} \left(\widetilde{Z} R \frac{\left[(\tau-T\wedge\tau)-e^{\delta^\top \widetilde{Z}}\mu_0(Z; \tau)\right]}{e^{\delta^\top \widetilde{Z}}\pi_1(Z)+\pi_0(Z)}\pi_0(Z) \right.\right.\\
    & \left.\left.~~~~~~~~~~~~~~~~~~~~~~~~~-\widetilde{Z}(1-R)\frac{\left[(\tau-T\wedge\tau)-\mu_0(Z; \tau)\right]e^{\delta^\top \widetilde{Z}}}{e^{\delta^\top \widetilde{Z}}\pi_1(Z_i)+\pi_0(Z)}\pi_1(Z)\right)\right]=0,
\end{align*}
where $K_{C^{(r)}} = P(C^{(r)} > t \mid Z=z)$.

In the second approach, we consider the score 
$\exp\{(\beta_1-\beta_0)^\top \widetilde{z}\}$, where $\beta_r$ is the root of the the equation
$$\E\left[\widetilde Z\left\{\E(\tau-T^{(r)}\wedge \tau \mid Z)- \exp(\beta_r^\top\widetilde{Z})\right\}\right]=0.$$
While this regression model is not always natural for arbitrary choices of $\tau$, it can still be a useful working model, for example, to approximately model the hazard ratio for rare events.
Luckily, like the Poisson regression model discussed above, this approach is robust to mis-specification of the model in the sense that it doesn't introduce spurious treatment effect heterogeneity.  Therefore, it remains a useful working model for estimating heterogeneous treatment effects measured by RMTL ratio.

The methods proposed in Sections 2.2.1 and 2.2.2 in the paper only need three minor modifications to be used for these estimands, as well. First, we need to change the outcome used in the regression analyses. Second, we need to develop an estimator $\what{\mu}_r(Z; \tau)$ that corrects for censoring. And third, we need to correct the estimating equations for censoring via inverse probability weighting, which necessitates an estimator of $K_{Cr}(t \mid z)$. The rest of this section details how to make these adjustments.

The first and simplest change is changing the outcome, from $Y^{(r)}$ to $\tau - T^{(r)} \wedge \tau$. Due to censoring, we do not always observe $T^{(r)}$, even when $R=r$. We will correct for this later when we do IPW adjustment in the third modification, so that the estimating equation only depends on the outcome $\tau - T^{(r)} \wedge \tau$ when it is observed.

The second change requires an initial estimator $\what{\mu}_r(z; \tau)$ based on any sensible regression models characterizing the relationship between $T_i^{(r)}$ and $Z_i.$  For example, one may employ the standard Cox model $\lambda_r(t \mid z)=\lambda_{r0}(t)\exp\{\eta^\top_r B(z)\}.$
Under this model,
$$\tau-\int_0^\tau \exp\left[-\what{\Lambda}_{r0}(t)\exp\{\what\eta_r^\top B(z)\}\right]dt$$
can be used to approximate $\E(\tau-T_i^{(r)}\wedge \tau|Z_i=z)$, 
where $B(z)$ is a set of basis functions, $\what{\Lambda}_{r0}(t)$ is the Breslow estimator for the cumulative baseline hazard function and $\what{\eta}_r$ is the maximum partial likelihood estimator for $\eta_r$ in arm $r$ \citep{cheng1998prediction}.
Alternatively, we may assume a regression model tailored for the restricted mean survival time \citep{tian2013predicting}:
$$\E(T^{(r)}\wedge\tau |Z=z)=\tau \frac{\exp\{\eta_r^\top B(z)\}}{1+\exp\{\eta_r^\top B(z)\}}$$
and estimate $\eta_r$ by solving the simple estimating equation
$$\sum_{i=1}^n I(R_i=r)\what{L}_i(r)B(Z_i)\left\{T_i\wedge \tau-\tau \frac{\exp\{\eta_r^\top B(Z_i)\}}{1+\exp\{\eta_r^\top B(Z_i)\}}\right\}=0,$$
where
\begin{equation}
    \what{L}_i(r) = \frac{I(T_i^{(r)} \wedge \tau<C_i^{(r)})}{\what{K}_{C^{(r)}}(T_i\wedge \tau \mid Z_i)},
    \label{eq:censor-weight}
\end{equation}
$I(T_i\wedge \tau<C_i)=\Delta_i+(1-\Delta_i)I(X_i\ge\tau),$ and $\what{K}_{Cr}(\cdot \mid z)$ is an estimator of the survival function of the censoring time $C_i^{(r)} \mid Z_i=z$ under a regression model, such as the Cox model, or a non-parametric survival model.
Under this model, $\E(\tau-T_i^{(r)}\wedge \tau|Z_i=z)$  can be approximated by 
$$ \frac{\tau}{1+\exp\{\what{\eta}_r^\top B(z)\}},$$
where $\what{\eta}_r$ is the root to the estimating equation above. Finally, random survival forest \cite{ishwaran2008random}, or other non-parametric survival estimators can be used, as well.

The third and final modification is to adjust the estimating equations for the two regressions or contrast regression to account for the fact that $T^{(r)}$ is sometimes censored. To do that, we can multiply the original $\what{W}_r(r)$ by $\what{L}_i(r)$ to get the updated weight,
$$\what{W}_i(r)= \what{L}_i(r) \left(r\frac{R_i}{\what{\pi}_1(Z_i)}+(1-r)\frac{1-R_i}{\what{\pi}_0(Z_i)}\right).$$
 The key assumption here is that the inverse probability weighting via $\what{L}_i(r)$ can correct the effect of right censoring, as long as $\what{K}_{Cr}(t\mid z)$ is a consistent estimator of the survival function of the censoring time $C^{(r)}$ given $Z=z$ and $P(C^{(r)}>\tau|Z=z)\ge \epsilon_\tau>0.$

Let $\what{\delta}$ be the estimator of $\delta_0$ in the first approach.  Similar to the count outcome, $\what{\delta}$ is a consistent estimator of $\delta_0$, if either the propensity score estimator $\what{\pi}_1(z)$ or the outcome predictions $ \what{\mu}_0^{(-k)}(z; \tau)$ is consistent. However, $\what{K}_{Cr}(z)$ must be consistent, either way, to appropriately adjust for censoring using the weights $\what{L}_i(R_i).$ Then, $D(z)$ is approximately $\exp\left(\what{\delta}^\top \widetilde{z}\right).$

With an estimator $\what{D}(z)$ of the CATE, we need to estimate 
$$ AD(c)=\frac{\E\{\tau-T^{(1)}\wedge\tau|\what{D}(Z)\ge c\}}{\E\{\tau-T^{(0)}\wedge \tau| \what{D}(Z)\ge c\}}$$
based on validation data. To this end, 
we may estimate the $AD(c)$ by constructing doubly robust estimators for
$\mu_1(c;\tau) = \E(\tau-T^{(r)}\wedge\tau |\what{D}(Z)\ge c), r=0, 1.$ This, again, requires an estimator $\what\mu_{cr}(z; \tau)$ of
$\mu_r(z, \tau)=\E(\tau-T^{(r)} \wedge \tau |Z=z)$, targeted at the distribution of samples with $\what{D}(z) > c$, which can be constructed by the approaches discussed above for estimating $\mu_{r}(z; \tau)$ when constructing the CATE estimate. Then,
$$\what{\mu}_1(c; \tau) = m_c^{-1}\sum_{\what{D}(Z_i^V)\ge c} \left[\what{\mu}_{cr}(Z_i^V, \tau)+\what{L}_{ci}^V(r)\what{W}_{ci}^V(r)\left\{(\tau-T_i^V\wedge\tau)-\what{\mu}_{cr}(Z_i^V, \tau)\right\}\right]$$
is a doubly robust estimator of $\E(\tau-T^{(r)}\wedge\tau |\what{D}(Z)\ge c)$, and
$$\frac{\what\mu_1(c; \tau)}{\what\mu_0(c; \tau)}$$ is a plug-in estimator of
$AD(c)$.

\section{Example of CATE vs ATE measured by HR} \label{section:exampleHR}

Consider the following example, where $\lambda_1(t|z)/\lambda_0(t|z)$ is constant, i.e., there is no treatment effect heterogeneity, but $\exp\{(\beta_1-\beta_0)^\top z\}$ depends on $z.$ Let 
$$\lambda_0(t \mid Z) = 2\{Z^{-1}I(t<1)+Z^{-2}I(1\le t<2)+Z^{-3}I(t\ge 2)\},$$ 
$\lambda_1(t \mid Z) = 0.5\lambda_0(t\mid Z),$ and  $Z\sim U(0, 2),$ where $I(\cdot)$ is the indicator function.
Solving the score equations from the mis-specified Cox model results in $(\beta_1, \beta_0)\approx (-1.85, -1.55)$ in the absence of any censoring, implying nontrivial treatment effect heterogeneity.

\section{Simulation Study for Time to Event Outcomes}
\begin{figure}[htpb!]
\centering
\caption{The first row: the ATE in subgroups of patients identified by different CATE scores (the proposed methods, naive regression and boosting) in two simulation settings and the ATE in subgroups of patients sorted according to the true $D(z)$ (gray solid curve); the second row: the distribution of correlation coefficients between true CATE and estimated CATE scores (after log-transformation). }
  \centering
  \includegraphics[scale=0.75]{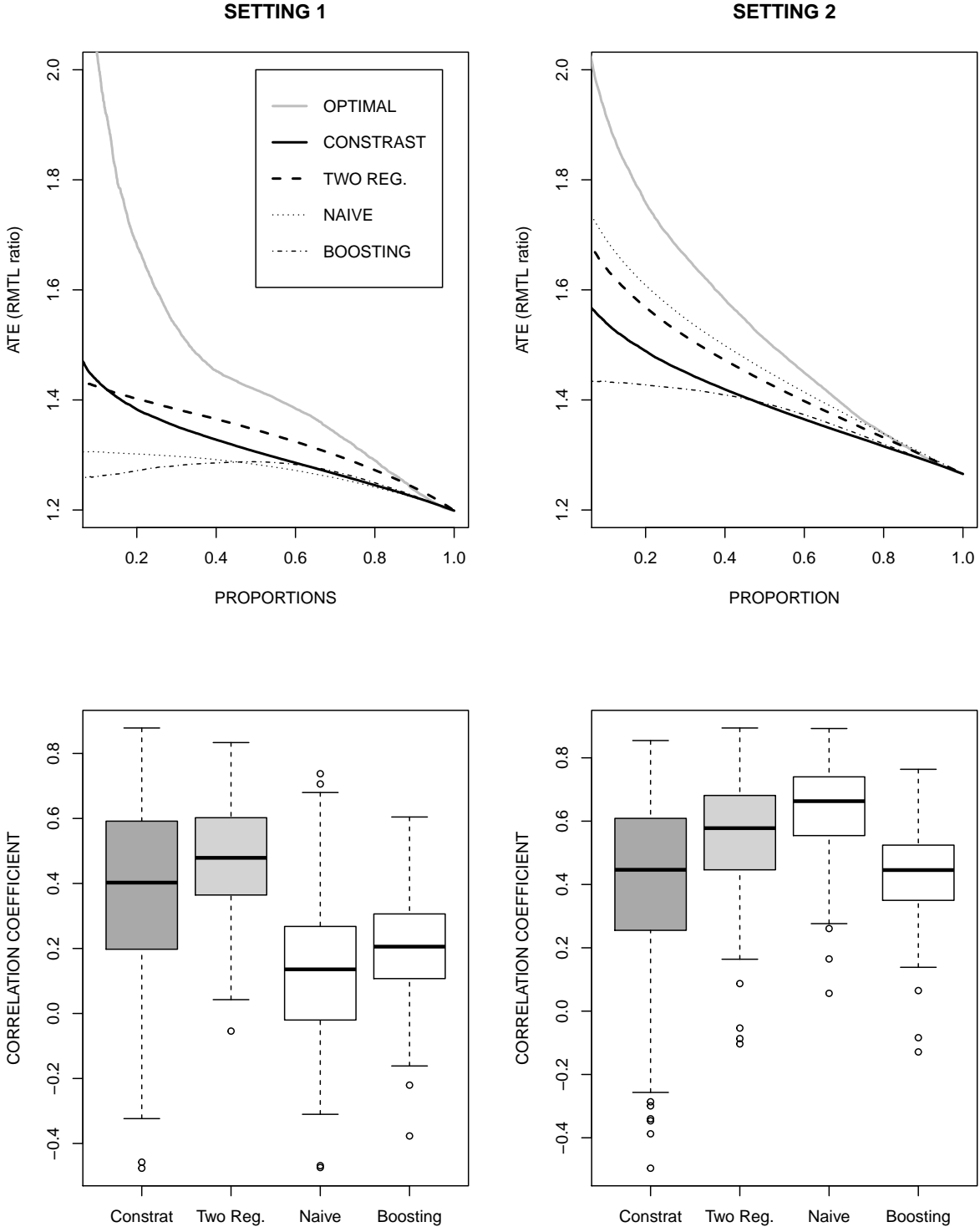}
  \label{fig:survsimulation}
\end{figure}

In this simulation study, we consider the survival outcomes:  $T^{(r)} \mid Z=z$ is generated from an exponential distribution with a failure rate  $\lambda_r(z)$.  The censoring distribution $C^{(r)} \mid Z=z$ is the minimum of a uniform random variable over [0.5, 1] and an exponential random variable with a failure rate of $\exp(0.25+z_3).$ The treatment assignment $R\mid z$ is a Bernoulli random variable with a probability of $\pi_1(z)=\left\{1+\exp(-z_1+0.5z_2+0.5z_6) \right\}^{-1}.$ Other than right censoring, a significant difference from the simulation study in the main paper is that the simple multiplicative model for $\mu_r(z)$ or $D(z)$ is always mis-specified.  Due to the anticipated similarity, we only consider the following two settings:
\begin{enumerate}
      \item The optimal CATE is approximately $\exp(-0.08+0.34z_1+0.11z_2+0.43z_6)$
      \begin{align*}
         \lambda_1(z)=&\exp\left(-0.350+0.125z_1+0.3z_2+0.5z_6-0.5|z_1|+0.5|z_2|\right)\\
         \lambda_0(z)=&\exp\left(-0.375+0.25z_2+0.25z_6-0.5|z_1|+0.5|z_2|\right)
    \end{align*}
    \item The optimal CATE is approximately $\exp(-0.09+0.27z_1+0.10z_2+0.51z_6)$ with
    \begin{align*}
        \lambda_1(z)=&\exp\left(-0.050+0.125z_1+0.3z_2+0.5z_6\right)\\
        \lambda_0(z)=&\exp\left(-0.075+0.25z_2+0.25z_6\right)
    \end{align*}
\end{enumerate}
In both settings, log-transformed true CATE is not equal to, but can be approximated by a linear combination of the covariates, which can explain approximately 85\% of the variation in log-transformed true CATE. $\tau_0=0.75$ in the RMTL and the censoring rate is 83\% in both settings.  In the light of the first set of simulations, we construct the CATE score only using four methods: (1) contrast regression; (2) two regression; (3) fitting a simple multiplicative regression model $\mu_r(z, \tau)=\exp(\beta_r^\top \widetilde{z})$ in each arm; and (4) the boosting method coupled with IPW correcting for right censoring to estimate $\mu_r(z, \tau)$ in each arm separately.  
The censoring probability in the IPW is estimated via a Cox regression model, which is misspecified.  Again, we calculate the the true validation curve based on constructed CATE scores and the correlation coefficients between the estimated CATE score and the true CATE after log-transformation.  After repeating this process 200 times, we summarize the performance of each method based on the average of the resulting validation curves and the distribution of correlation coefficients between the estimated and true CATEs in Figure \ref{fig:survsimulation}.

In the first setting, the two proposed methods generate the best CATE score based on either the validation curve or the  distribution of correlation coefficients.  The estimated CATE score based on na\"ive regression or boosting is not informative. In the second setting, where the multiplicative regression model is approximately correct within each arm, the na\"ive regression approach performs the best, as expected. The two regression performs almost equally well. Both boosting and contrast regression also yield informative CATE scores for subgroup identification. In general, the simulation results are consistent with those in the main paper.

\section{NDT Example for Time to the First Relapse}

In the analysis of the time to the first relapse, we use the ratio of RMTL due to relapse as the metric for the treatment effect, with a truncation time point $\tau_0$ of 4.34 years (52 months). In the na\"ive comparison without adjusting for baseline covariates, the estimated ratio of mean time lost up to $4.34$ years is 1.246 (95\% confidence interval: 1.098-1.414, $p<0.001$), suggesting that time lost to relapse in patients receiving DMF is only 80\% of that in patients receiving TERI. After the doubly robust adjustment, the estimated ratio becomes 1.531 (95\% confidence interval: 1.313-1.785, $p<0.001$) with the RMTL of 1.451 and 0.947 years for TERI and DMF, respectively, showing a greater difference between two treatments. We compare CATE scores based on the na\"ive regression and new proposal. In constructing the proposed CATE scores, we used random forest with 50 trees to generate the initial prediction of the RMTL due to relapse. The log-transformed CATE score is a linear combination of baseline covariates, whose composition is also reported in Table \ref{table:survscore} of the Supplementary Material. While the contributions of the same covariate to resulting CATE scores can be different, such as the EDSS score at baseline, the resulting CATE scores are correlated.  We also performed cross validation to evaluate the performance of the different CATE scores. It appears that the CATE score directly based on random forest overfits the training data, but performs similarly as other methods in the testing set. The estimated treatment effect heterogeneity is very weak based on any of the four CATE scores. For example,  in the cross-validation, the estimated ratio of RMTL in 50\% of the patients favoring DMF based on contrast regression approach is 1.624. The ratio in the rest of the 50\% patients is 1.460, only marginally lower. The distributions of two RMTL ratios across replicates substantially overlap. Lastly, we examine the relationship between the CATE scores built for the number of relapses and the time to the first relapse. The two sets of CATE scores are correlated as expected. For example the correlation coefficient between two CATE scores based on the contrast regression is 0.80.

\section{NTD Registry Data Management}
All data are pseudonymized and pooled to form the MS registry database. The codes are managed by the Institute for medical information processing, biometry and epidemiology (Institut für medizinische Informationsverarbeitung, Biometrie und Epidemiologie) at the Ludwig Maximilian University in Munich, Germany, acting as an external trust center. 
This data acquisition and management protocol was approved by the ethical committee of the Bavarian Medical Board (Bayerische Landesärztekammer; June 14, 2012) and re-approved by the ethical committee of the Medical Board North-Rhine. (Ärztekammer Nordrhein, April 25, 2017). 
Compliance with European and German legislation 
is warranted including patient rights and informed consent requirements.  The data accuracy is ensured by dynamic web-based data capturing, regular training of doctors and nurses, interactive chat forum for nurses and doctors, automated and manual feedback query system, daily automated analysis of data plausibility and correctness, and annually on-site audit of procedures and source data by an external process quality certifier organization.

\newpage
\begin{table}[!htbp]
    \caption{The estimated weights in constructed CATE scores (TERI vs DMF) }
    \label{table:survscore}
      \centering
      \begin{tabular}{c rrr }
  \hline\hline
 & \multicolumn{3}{c}{ratio of RMTL}\\
  \cline{2-4} 
   & na\"ive reg. & two reg. & contrast reg.\\
 \hline
intercept                     &  1.061 & 1.310 & 1.017\\
age                           &  0.008 & 0.011 & 0.010\\
\# prior treatments           & -0.255 &-0.342 &-0.210\\
MS duration                   &  0.004 & 0.003 & 0.007\\
(years)                       &&&\\       
GA                            & -0.630 &-0.801 &-0.640\\
IFN                           & -0.390 &-0.583 &-0.410\\
\# relapses                   & -0.217 &-0.193 &-0.126\\
(prior year)                  &&&\\
\# relapses                   &  0.195 & 0.139 & 0.130\\
(prior two years)             &&&\\
EDSS                          & -0.072 &-0.060 &-0.117\\
pyramidal EDSS                &  0.060 & 0.030 & 0.073\\
\hline
\end{tabular}
\end{table}

\newpage

\newpage

\newpage
\begin{figure}[htpb!]
\centering
\caption{The ATE measured by the RMTL ratio in subgroups of patients based on the CATE scores constructed in the training set (two proposed methods, naive regression and random forest) in the NTD registry}
  \centering
  \includegraphics[scale =0.6]{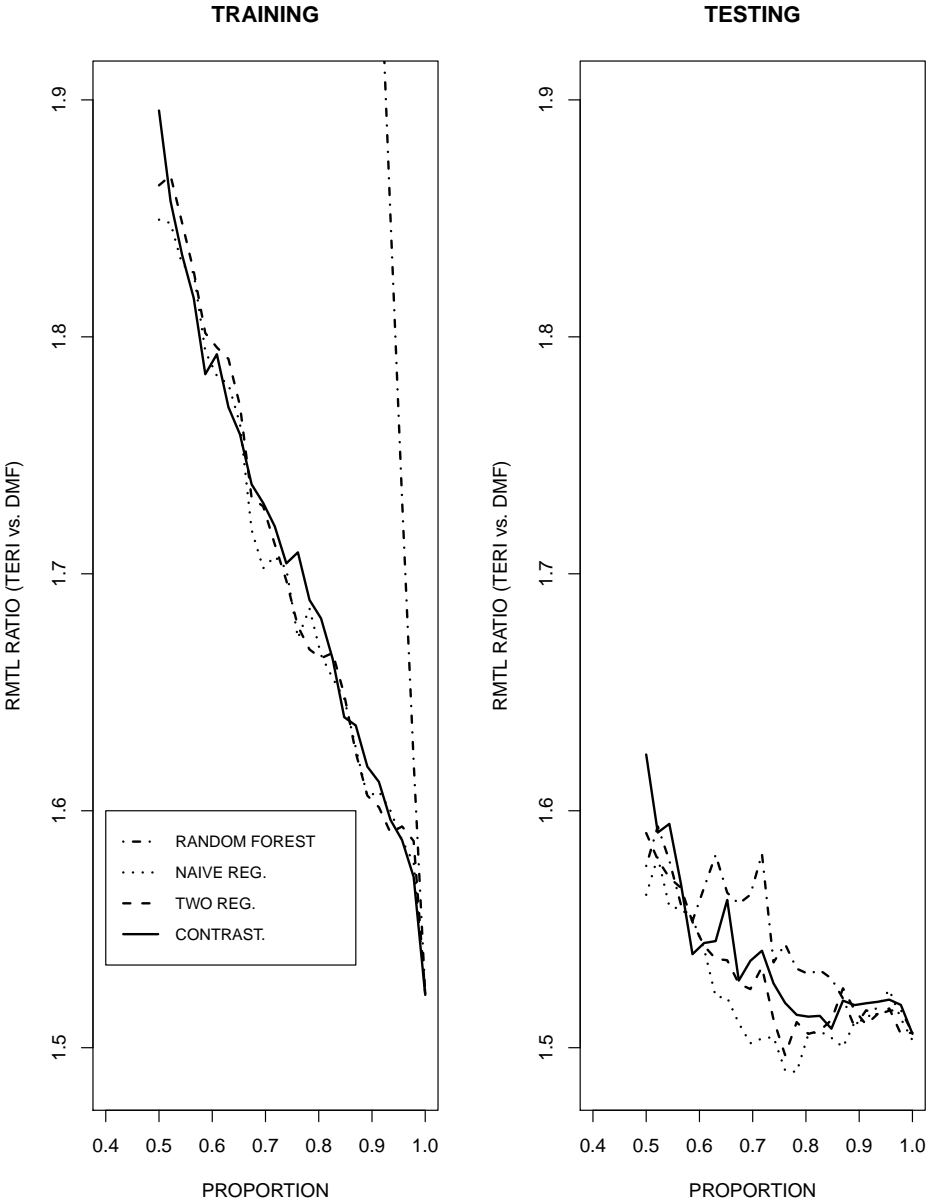}  
  \label{fig:supp-exp1}
\end{figure}

\newpage
\begin{figure}[htpb!]
\centering
\caption{The cross-validated ATE (RMTL ratio of TERI vs DMF) of subgroups of patients identified by different CATE scores ( two proposed method, naive regression and random forest) in the NTD registry.}
  \centering
  \includegraphics[scale=0.6]{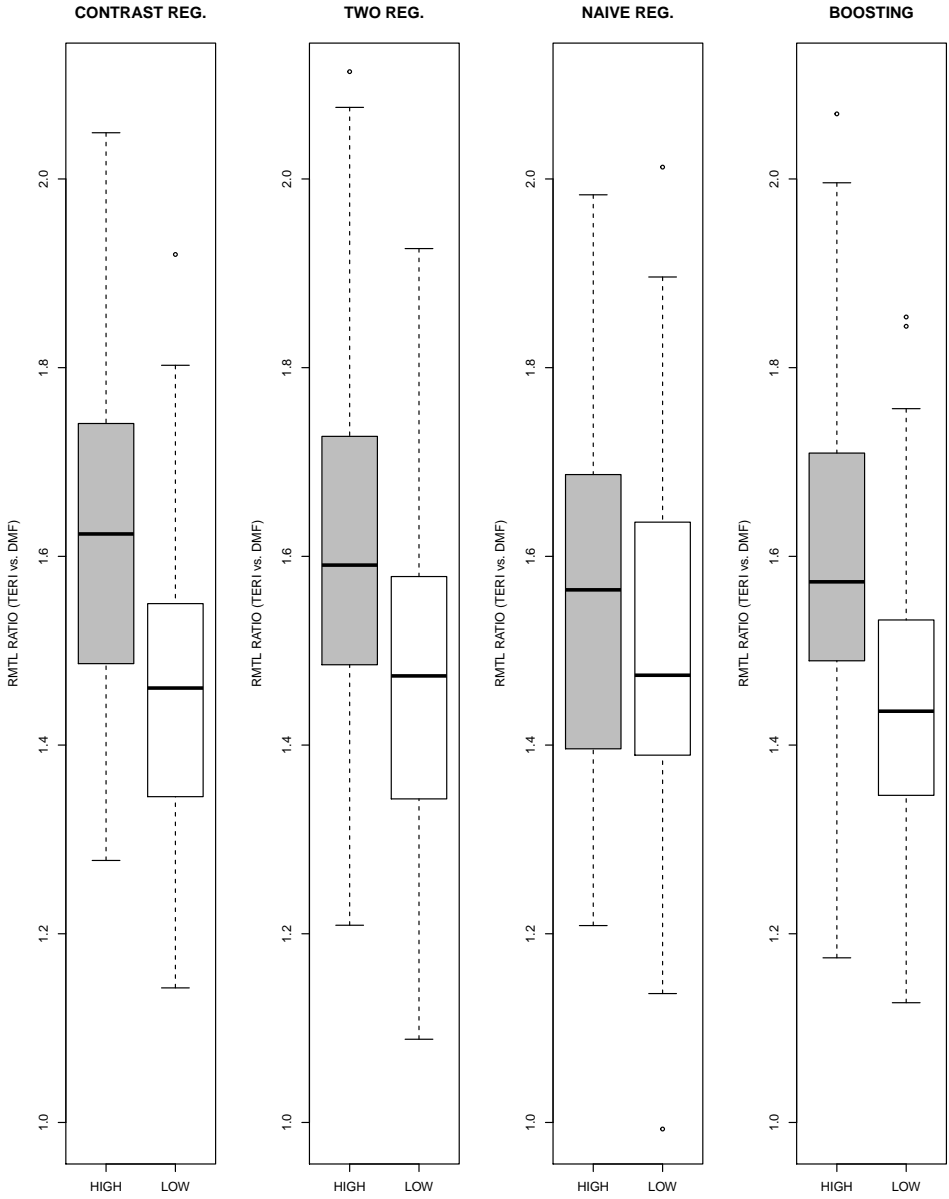}
  \label{fig:supp-exp2}
\end{figure}

\end{document}